\documentclass[letterpaper]{article} 
\usepackage{aaai2026}  
\usepackage{times}  
\usepackage{helvet}  
\usepackage{courier}  
\usepackage[hyphens]{url}  
\usepackage{graphicx} 
\usepackage{natbib}  
\usepackage{caption} 
\usepackage{tikz}
\usepackage{amsfonts}
\usepackage{amsmath}
\usepackage{amssymb}
\usepackage{amsthm}
\usepackage{bm}
\usepackage{booktabs}
\usepackage{comment}

\usepackage{multirow} 

\usepackage{tabularx} 
\usepackage{array, colortbl} 

\newtheorem{definition}{Definition}

\usepackage[table,xcdraw]{xcolor}

\definecolor{TablePurple}{RGB}{255, 255, 255}
\definecolor{TableBlue1}{RGB}{255, 255, 255}   
\definecolor{TableBlue2}{RGB}{255, 255, 255}  
\definecolor{TableGreen}{RGB}{255, 255, 255} 
\definecolor{TableYellow}{RGB}{255, 255, 255} 

\usepackage{algorithm}
\usepackage{algorithmicx}
\usepackage{algpseudocode}
\usepackage[switch]{lineno}

\newcommand{\projname}{{3D-ANC}}

\usepackage{newfloat}
\usepackage{listings}
\DeclareCaptionStyle{ruled}{labelfont=normalfont,labelsep=colon,strut=off} 
\floatstyle{ruled}
\newfloat{listing}{tb}{lst}{}
\floatname{listing}{Listing}
\title{\emph{\projname}: Adaptive Neural Collapse for Robust 3D Point Cloud Recognition}
\author {
    Yuanmin Huang\textsuperscript{\rm 1},
    Wenxuan Li\textsuperscript{\rm 1},
    Mi Zhang\textsuperscript{\rm 1}\footnotemark[2],
    Xiaohan Zhang\textsuperscript{\rm 1},
    Xiaoyu You\textsuperscript{\rm 2},
    Min Yang\textsuperscript{\rm 1}\thanks{Corresponding authors: Mi Zhang and Min Yang.}
}
\affiliations {
    \textsuperscript{\rm 1}College of Computer Science and Artificial Intelligence, Fudan University, Shanghai, China\\
    \textsuperscript{\rm 2}School of Information Science and Engineering, East China University of Science and Technology, Shanghai, China\\
    yuanminhuang23@m.fudan.edu.cn, 
    wxli24@m.fudan.edu.cn,
    mi\_zhang@fudan.edu.cn,
    xh\_zhang@fudan.edu.cn,
    xiaoyuyou@ecust.edu.cn,
    m\_yang@fudan.edu.cn
}

\begin{document}

\maketitle

\begin{abstract}
Deep neural networks have recently achieved notable progress in 3D point cloud recognition, yet their vulnerability to adversarial perturbations poses critical security challenges in practical deployments. 
Conventional defense mechanisms struggle to address the evolving landscape of multifaceted attack patterns. 
Through systematic analysis of existing defenses, we identify that their unsatisfactory performance primarily originates from an entangled feature space, where adversarial attacks can be performed easily.
To this end, we present \emph{\projname}, a novel approach that capitalizes on the Neural Collapse (NC) mechanism to orchestrate discriminative feature learning. 
In particular, NC depicts where last-layer features and classifier weights jointly evolve into a simplex equiangular tight frame (ETF) arrangement, establishing maximally separable class prototypes. 
However, leveraging this advantage in 3D recognition confronts two substantial challenges: (1) prevalent class imbalance in point cloud datasets, and (2) complex geometric similarities between object categories.
To tackle these obstacles, our solution combines an ETF-aligned classification module with an adaptive training framework consisting of representation-balanced learning (RBL) and dynamic feature direction loss (FDL). 
\emph{\projname} seamlessly empowers existing models to develop disentangled feature spaces despite the complexity in 3D data distribution. 
Comprehensive evaluations state that \emph{\projname} significantly improves the robustness of models with various structures on two datasets. For instance, DGCNN's classification accuracy is elevated from $27.2\%$ to $80.9\%$ on ModelNet40 -- a $53.7\%$ absolute gain that surpasses leading baselines by $34.0\%$.
\end{abstract}


\section{Introduction}
Point cloud data, which captures the 3D structures of objects and environments, is increasingly crucial in various applications, including autonomous driving, robotics, and 3D scene understanding \cite{hamdi2021mvtn, qiu2021geometric, abbasi2022lidar, duan2021robotics}. 
In recent years, deep neural networks (DNNs) have achieved remarkable success in point cloud recognition, largely due to their deep, non-linear architectures \cite{qi2017pointnet,qi2017pointnet++,wang2019dynamic}. 
However, alongside their widespread adoption, there has been growing concern over their vulnerability to \emph{adversarial attacks}. 
As an attacker, one can perform imperceptible modifications to point clouds to drastically alter the predictions of a DNN. 
Given the significant risks posed to real-world applications, enhancing the adversarial robustness of point cloud recognition has emerged as a critical research area in 3D vision. 

\begin{figure}[t]
    \centering
    \includegraphics[width=0.9\linewidth]{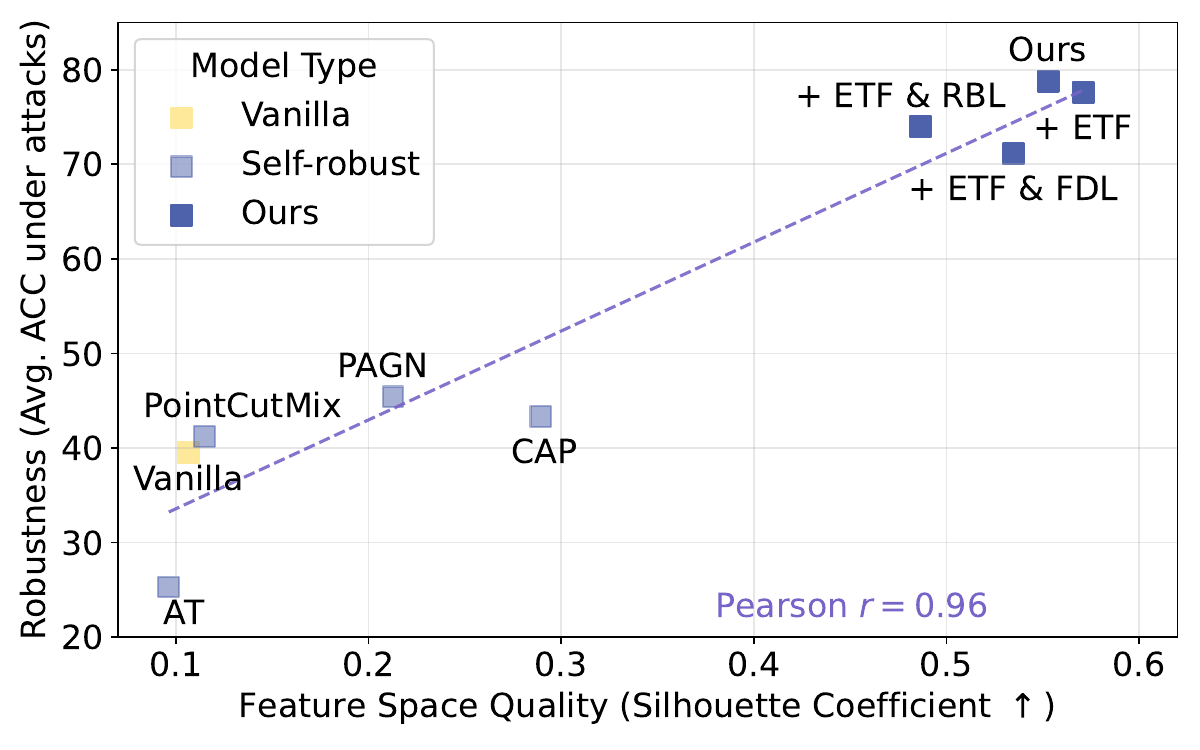}
    \caption{Adversarial robustness highly correlates with feature space quality. However, current models and self-robust defenses have poor feature disentanglement ability. {\projname} significantly improves robustness with superior feature separability. Experiments are conducted on ModelNet40, PointNet. Detailed setting is described in Sec. \ref{sec:feat_qlty}. }
    \label{fig:feat_qlty}
\end{figure}   

Since point cloud data consists of 3D coordinates of discrete points, this unique data format provides attackers with various perturbation strategies. 
Typical adversarial attacks for point clouds involve adding \cite{xiang2019generating}, deleting \cite{zheng2019pointcloudsaliency}, or shifting \cite{kim2021minimal} points (i.e., altering the positions of existing points) within a point cloud. 
Additionally, attackers may employ generative models to transform a benign point cloud into an adversarial one \cite{hamdi2020advpc,zhou2020lggan}. 
More recent shape-invariant attacks \cite{huang2022shape,lou2024hide} preserve the geometric structure of benign samples, thereby generating adversarial examples hard to detect.
Given the diversity and sophistication of these attacks, the demand for robust defense mechanisms in point cloud recognition has become increasingly urgent. 

Current defenses can be broadly classified into \emph{input preprocessing} \cite{zhou2019dupnet,sun2023critical} and \emph{self-robust models} \cite{liu2019extending,sun2021adversarially}. 
Input preprocessing focuses on mitigating anomaly patterns introduced by certain attacks, such as outlier points in a point cloud. 
While effective, this approach is limited in its ability to counter more advanced attacks, e.g., shape-invariant attacks \cite{lou2024hide}, which tend to produce fewer outliers. 
In contrast, self-robust models aim to improve the model's intrinsic robustness against potential adversarial threats. 
For example, techniques such as adversarial training \cite{Zhang2019DefenseAA} or self-supervised learning \cite{sun2021adversarially,zhang2022pointcutmixa} can be employed to develop robust models. 
Nonetheless, our pilot study (Sec. \ref{sec:preliminary_results}) reveals that existing self-robust models often struggle to extract well-separated sample features.
This can lead to unsatisfying robustness under adversarial conditions, as benign samples can be easily perturbed across decision boundaries when the feature space overlaps between classes. 

\textit{Faced with emerging attacks, a key challenge in developing effective self-robust models is to achieve a robustly disentangled feature space.} 
To tackle this issue, we draw inspiration from the recently discovered \textit{Neural Collapse (NC)} phenomenon \cite{papyan2020prevalence}. 
It has been observed that, during the terminal phase of training on a balanced dataset, the last-layer features of same-class samples tend to collapse to their within-class mean. 
Simultaneously, the within-class means of all classes, along with the corresponding classifier vectors, converge to the vertices of a simplex equiangular tight frame (ETF).
An ETF structure is depicted in Fig. \ref{fig:framework}(a), which maximizes the pairwise angular separation between vectors. 
As a result, NC offers a promising solution to a well-disentangled feature space: replacing the traditional learnable classification head with a non-learnable one satisfying ETF structure. 
This should enhance the adversarial robustness of a model inherently.

However, adapting NC to point cloud recognition differs drastically from image classification under typical balanced settings. 
Specifically, two substantial challenges arise: \textit{(1) the imbalanced nature of point cloud datasets}, which can severely distort the training of models with ETF classifiers, degrading performance on clean samples; and \textit{(2) the complex inter-class geometric similarities}, where samples from distinct classes can share nearly identical geometries, causing feature overlap even when trained with ETF classifiers, leading to suboptimal robustness performance. 

To address these challenges, we propose \textbf{\projname}, an \textbf{\underline{a}daptive} training framework designed to facilitate the integration of \underline{\textbf{NC}} (an ETF classifier) into \underline{\textbf{3D}} point cloud recognition.
The framework comprises two key components: a \textit{representation-balanced learning (RBL)} module and a \textit{dynamic feature direction loss (FDL)}.
Built upon an ETF classifier, RBL introduces learnable rotations to the ETF head, enabling adaptation to imbalanced class distributions while preserving ETF properties.
Meanwhile, the dynamic FDL reinforces both intra-class compactness and inter-class repulsion, adaptively targeting geometrically similar classes that hinder feature disentanglement. 
Overall, {\projname} achieves a robustly disentangled feature space tailored to point cloud data.

Our main contributions are summarized as follows:
\begin{itemize}
    \item We introduce {\projname}, the first approach to leverage the NC phenomenon for robust point cloud recognition. 
    By incorporating an ETF classifier within an adaptive training framework, {\projname} inherently preserves class-specific features and maximizes inter-class repulsion, significantly improving adversarial robustness.
    \item {\projname} is model-agnostic and computationally efficient, seamlessly integrating with existing point cloud recognition models across architectures by simply replacing the classification head with an ETF one.
    \item {\projname} achieves substantial improvements in adversarial robustness across three widely used models on two benchmark datasets. For example, {\projname} boosts the classification accuracy of DGCNN from $27.2\%$ to $80.9\%$, outperforming the best baseline, which attains only $46.9\%$.
\end{itemize}

\section{Related Work}

\subsection{Point Cloud Recognition}
Point cloud recognition involves the task of categorizing objects represented as sets of points with 3D coordinates into distinct classes. 
This problem can be formally described as learning a mapping function $F: \mathbf{X} \mapsto y$, where $\mathbf{X} \in \mathbb{R}^{N \times 3}$ represents a point cloud composed of $N$ points in 3D space, and $y \in \{1, 2, \dots, C\}$ denotes the ground truth label among $C$ possible categories. 
In recent years, DNNs have shown great success in point cloud recognition due to their ability to extract complex, non-linear features \cite{lu2020deep}. 
This paper focuses on point-based models \cite{qi2017pointnet,qi2017pointnet++}.
Unlike approaches that require pre-processing steps such as voxelization \cite{xie2020voxelnet} or pillarization \cite{lang2019pointpillars}, point-based models directly take the 3D coordinates of point clouds as input. 
These models start by extracting latent features from each of the $N$ points in the input point cloud $\mathbf{X} \in \mathbb{R}^{N \times 3}$ using a non-linear feature extractor $f$, such as an MLP in PointNet/PointNet++ \cite{qi2017pointnet,qi2017pointnet++}, a GNN in DGCNN \cite{wang2019dynamic}, a CNN in PointCNN \cite{li2018pointcnn}, or a Transformer in PCT \cite{guo2021pct}. 
The extracted global feature $\mathbf{h}$ is then passed through a fully-connected layer that serves as the classification head $g$. For a more detailed review, please refer to \cite{lu2020deep}.

\begin{figure*}[t]
    \centering
    \includegraphics[width=0.9\linewidth]{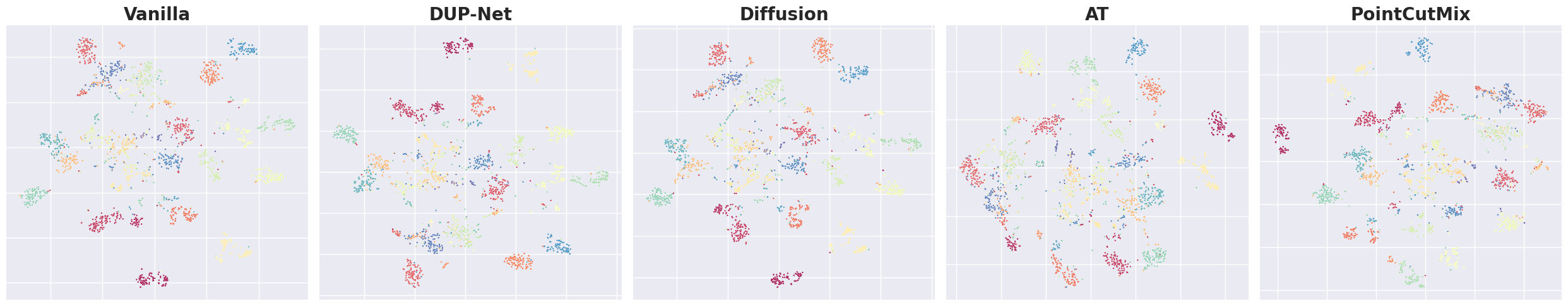}
    \caption{The t-SNE visualization of PointNet features on ModelNet40 test set under various defense schemes. For input preprocessing defense (DUP-Net and Diffusion), we use the feature of the preprocessed samples for visualization. The entangled feature space leaves room for adversarial attacks, i.e., samples can be perturbed to overlapped classes easily. }
    \label{fig:pilot_tsne}
\end{figure*}

\subsection{Adversarial Examples}
Adversarial attacks are designed to manipulate the output of DNN models by introducing small perturbations to the input data. 
Following attacks on image classification models \cite{szegedy2014intriguing,goodfellow2014explaining}, adversarial attacks also appear in point cloud recognition \cite{xiang2019generating,wen2020geometry}.
The objective can be formulated as:
\begin{equation} \label{eq:atk_goal}
    \max_\mathbf{\eta} \mathcal{L}(F(\mathbf{\tilde{X}}, y)), \text{\quad s.t.}, ||\mathbf{\eta}||_p \le \delta, 
\end{equation}
where $F$ represents the model, $\mathcal{L}$ is the classification loss function, $\mathbf{\eta}$, $\mathbf{\Tilde{X}} = \mathbf{X} \oplus \mathbf{\eta}$, and $y$ denote the perturbation, the adversarial example, and the ground truth label, respectively. 
Adversarial attacks in point cloud classification exhibit a wide range of strategies. 
With respect to the form of perturbations (i.e., $\oplus$ in Eq.\ref{eq:atk_goal}), options include shifting \cite{xiang2019generating}, adding \cite{liu2020adversarialstick,xiang2019generating} or deleting \cite{wicker2019robustnesskey,zheng2019pointcloudsaliency} points, or transforming the entire point cloud \cite{hamdi2020advpc,zhou2020lggan}. 
Regarding the perturbation budget $||\mathbf{\eta}||_p$, various distance metrics have been employed, such as $L_0$, $L_2$, and $L_\infty$, with Chamfer Distance often used as a representative for $L_2$ distance \cite{xiang2019generating}.
Recently, shape-invariant attacks \cite{huang2022shape,wen2020geometry} have been proposed to maintain local or global geometric patterns of the point cloud during attacks, resulting in more imperceptible perturbations.

Adversarial defenses for point cloud classifiers have evolved in tandem with the development of attacks. 
Recent approaches can be broadly divided into input preprocessing \cite{zhou2019dupnet,sun2023critical} and self-robust models \cite{liu2019extending,Zhang2019DefenseAA,sun2021adversarially}. 
Input preprocessing methods mitigate specific effects introduced by particular attacks. 
For instance, SOR \cite{zhou2019dupnet} filters out outlier points introduced by adversarial perturbations, while DUP-Net \cite{zhou2019dupnet} addresses inconsistent point density within a point cloud caused by such attacks. 
PointDP \cite{sun2023critical} employs diffusion models to purify the perturbed point cloud during the diffusion process. 
Conversely, self-robust models draw inspiration from defenses in image classification. 
For example, adversarial training methods \cite{madry2017towards} augment the training dataset with adversarial examples. 
Additionally, some models \cite{sun2021adversarially,zhang2022pointcutmixa,ding2023cap} utilize self-supervised learning to enhance the classifiers in the presence of noisy inputs, thereby improving adversarial robustness.

Despite their effectiveness, both defense strategies share a common limitation: \emph{they generalize poorly to unseen attacks with characteristics different from those they were designed to counter. }
When confronted with novel attacks, input preprocessing techniques may fail to remove malicious points, and existing self-robust models may remain vulnerable. 
In contrast, our approach focuses on constructing a naturally well-disentangled feature space for point cloud recognition models. 
This design inherently equips our method with the capability to resist a wide range of attacks.

\subsection{Neural Collapse}

Neural Collapse (NC) was first introduced in \cite{papyan2020prevalence}, describing how, in a well-trained classification model, the last-layer features collapse to their within-class centers. 
This collapse forms a simplex equiangular tight frame (ETF) structure, where the feature vectors align with the classifier’s weight vectors. 
Subsequent research has demonstrated that NC represents global optimality under balanced training with both cross-entropy (CE) loss \cite{fang2021exploring, DBLP:conf/iclr/JiLZDS22, lu2022neural} and mean squared error (MSE) loss \cite{DBLP:conf/iclr/HanPD22, tirer2022extended, zhou2022optimization}.

Further studies \cite{yang2022inducing, peifeng2023feature, zhong2023understanding, thrampoulidis2022imbalance, xie2023neural} have examined NC in the context of imbalanced training. 
Yang et al. \cite{yang2022inducing} showed that using a fixed ETF classifier head can align features with the ETF structure, improving classification performance even with imbalanced datasets. 
Zhong et al. \cite{zhong2023understanding} observed that while imbalanced distributions can disrupt the NC structure, initializing the classifier as a simplex ETF still benefits minor class differentiation. 

Inspired by these findings, we aim to induce an equiangular separation among class centers to enhance feature disentanglement in the representation space, thereby improving the model’s robustness against adversarial attacks.

\section{Pilot Study}

\subsection{Preliminary Results} \label{sec:preliminary_results}
We first conduct preliminary experiments to evaluate existing defenses. 
Given that adversarial attacks introduce subtle perturbations to point clouds—imperceptible to humans yet capable of deceiving model predictions—we hypothesize that the vulnerabilities of current defenses may originate from weaknesses in the feature space.
To examine this hypothesis, we analyze the feature extraction capabilities of existing models and defenses, as illustrated in Fig. \ref{fig:pilot_tsne}. 
The models are evaluated with the output of the feature extractor $h = f(X)$ as the feature given benign samples. 

As depicted in the figure, the vanilla PointNet faces challenges in distinguishing features across different classes. 
Although input preprocessing techniques (e.g., DUP-Net \cite{zhou2019dupnet}, Diffusion \cite{sun2023critical}) and existing self-robust models (e.g., Adversarial Training \cite{liu2019extending}, PointCutMix \cite{zhang2022pointcutmixa}) offer some enhancement, \emph{substantial overlap remains among inter-class features.}
More visualizations of advanced models in Appendix \ref{sec:more_defense_visual} exhibit similar trends. 
Now that \emph{point cloud models struggle to extract well-separated features even from clean samples, adversarial examples can easily go across decision boundaries, posing considerable risk}.
Such an observation motivates us to improve the feature extraction capabilities of existing point cloud recognition models.

\subsection{Neural Collapse Phenomenon}
\begin{figure*}[t]
    \centering
    \includegraphics[width=0.92\linewidth]{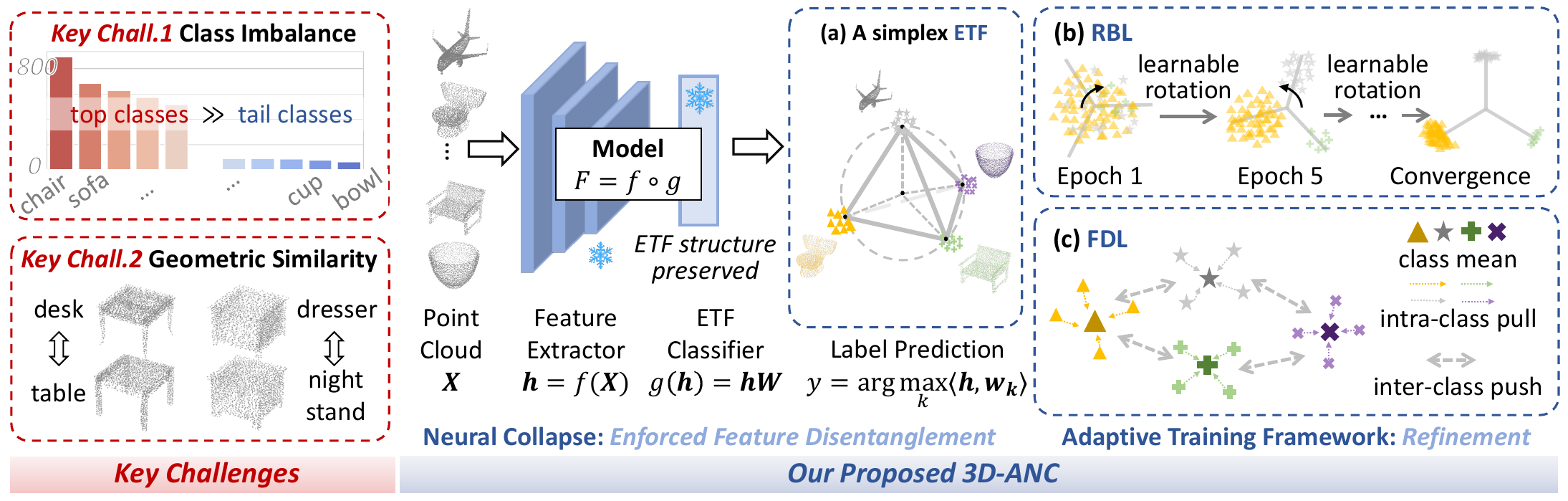}
    \caption{
        \textbf{Overview of the proposed {\projname}.} 
        \textbf{(Left)}: Key challenges in point cloud recognition to form a well-separated feature space: (1) class imbalance and (2) geometric similarity between certain classes. 
        \textbf{(Right)}: (a) Our {\projname} enforces feature disentanglement via ETF classifier $g$. 
        To perform fine-grained refinement towards feature separation for robust point cloud recognition, we further propose an adaptive training framework comprising (b) Representation-Balanced Learning and (c) dynamic Feature Direction Loss, which progressively adapts the training with ETF classifier.
    }
    \label{fig:framework}
\end{figure*}

We then introduce the Neural Collapse (NC) phenomenon. 
As explored in \cite{papyan2020prevalence}, for neural networks trained on a balanced dataset, the NC phenomenon occurs in the last-layer features and the weight vectors of the classification head as training converges. 
Specifically, both the features and weight vectors collapse to the vertices of a simplex equiangular tight frame (ETF), as depicted in Fig. \ref{fig:framework}(a). 
Formally, a simplex ETF is defined as follows:

\begin{definition}[Simplex Equiangular Tight Frame]
    \label{def:etf}
    A collection of vectors $\mathbf{m}_i \in \mathbb{R}^{d}, i = 1, 2, \cdots, K, d \ge K-1$, is said to be a simplex equiangular tight frame if:
    \begin{equation} \label{eq:etf}
        \mathbf{M} = \sqrt{\frac{K}{K-1}} \mathbf{R} \left(\mathbf{I}_K - \frac{1}{K} \mathbf{1}_K \mathbf{1}_K^T\right),
    \end{equation}
    where $\mathbf{M} = \left[\mathbf{m}_1, \cdots, \mathbf{m}_K \right] \in \mathbb{R}^{d \times K}$, $\mathbf{R} \in \mathbb{R}^{d \times K}$ represents a rotation matrix satisfying $\mathbf{R}^T\mathbf{R}=\mathbf{I}_K$, $\mathbf{I}_K$ is the identity matrix, and $\mathbf{1}_K$ is a vector of all ones. 
\end{definition}

This definition implies that a simplex ETF has the property where any pair of vectors exhibits the same maximal angular separation, i.e., $\mathbf{m}_i^T \mathbf{m}_j = -\frac{1}{K-1}$ for $i \neq j$, where $i,j \in {1, \cdots, K}$.

We now formally describe the NC phenomenon:

\noindent\textbf{NC1: Variability collapse.} 
The last-layer feature $\mathbf{h}_{k, i}$ of any sample $i$ from class $k$ collapses to the within-class mean $\mathbf{\bar{h}}_k = \text{Avg}_i \{\mathbf{h}_{k, i}\}$. 
As a result, the within-class feature variability collapses to zero.

\noindent\textbf{NC2: Convergence to simplex ETF.} 
The normalized within-class means converge to a simplex ETF that satisfies Eq. \ref{eq:etf}, i.e., $\mathbf{\tilde{h}}_k = \left(\mathbf{\bar{h}}_k - \mathbf{\bar{h}}_G \right) / ||\mathbf{\bar{h}}_k - \mathbf{\bar{h}}_G ||$ for $k\in \{1,K\}$, where $\mathbf{\bar{h}}_G$ represents the global mean across all classes.

\noindent\textbf{NC3: Self-duality.}
Classifier weight $\mathbf{w}_k$ for class $k$ aligns with the normalized class mean, i.e., $\mathbf{\tilde{h}}_k = \mathbf{w}_k / || \mathbf{w}_k ||$.

\noindent\textbf{NC4: Nearest class center prediction.}
The model's prediction for any sample feature $\mathbf{h}$ collapses to selecting the nearest class mean (center), i.e., $\text{argmax}_k \langle \mathbf{h},\mathbf{w}_k \rangle = \text{argmin}_k ||\mathbf{h}-\mathbf{\bar{h}}_k||$.

\section{The Proposed {\projname}}

\subsection{Neural Collapse for Robust Recognition} \label{sec:method_etf}
Preliminary results in Sec. \ref{sec:preliminary_results} highlight the inadequacies of current models and defenses in effectively extracting discriminative features. 
Fortunately, the NC phenomenon offers an ideal framework for achieving a well-disentangled feature space, characterized by highly cohesive intra-class features (NC1) and distinct inter-class features (NC2).

Instead of allowing the model to gradually converge to NC during training, we propose to initialize the model's classification head with a predefined neural collapse state, specifically a simplex Equiangular Tight Frame (ETF) head (Fig. \ref{fig:framework} (a)). 
Formally, for a model $F = f \circ g$ consisting of a feature extractor $f(X)=\mathbf{h}$ and a classifier $g(\mathbf{h})$, we set $g(\mathbf{h})=\mathbf{h}\cdot \mathbf{W}$, where $\mathbf{W}=\left[\mathbf{w}_1, \cdots, \mathbf{w}_K\right]$ represents a randomly initialized simplex ETF, as defined in Eq. \ref{eq:etf}.

According to NC3 and NC4, in an NC-convergent model, the feature $\mathbf{h}_k$ of class $k$ aligns with both the classification vector $\mathbf{w}_k$ and the normalized class mean $\mathbf{\tilde{h}}_k$. 
To encourage this alignment, we introduce a dot product loss defined as:
\begin{equation}
    \mathcal{L}_{dot}(\mathbf{h}, \mathbf{W}) = \frac{1}{2\sqrt{E_W E_H}} \left(\mathbf{w}_k^T \mathbf{h} - \sqrt{E_W E_H}\right)^2 ,
\end{equation}
where $\mathbf{h}$ denotes the feature of a sample belonging to class $k$, and $\mathbf{w}_k$ is the classification vector for class $k$. 
The terms $E_W$ and $E_H$ are predefined $L_2$-norm constraints, ensuring that $||\mathbf{h}||^2 \le E_H$ and $||\mathbf{w}_k||^2 \le E_W$.

To guarantee a thoroughly disentangled feature space, the matrix $\mathbf{W}$ remains fixed during training. 
By enforcing the simplex ETF head, the feature extractor is compelled to learn features that inherently align with the classifier vector $\mathbf{w}_k$ for each class $k$.
This design encourages the feature extractor to focus on class-specific characteristics within each point cloud, thereby enhancing the model's robustness.

\subsection{Adaptive Training Framework} 
While Yang et al. \cite{yang2022inducing} suggested that directly training a model with a fixed simplex ETF head can lead to effective convergence in image classification, we find this approach challenging to implement for point cloud data. 

Unlike image datasets, point cloud datasets often exhibit complex distribution characteristics that hinder direct application of ETF-based training.
For instance, the widely used ModelNet40 dataset \cite{wu20153dshapenet} \textit{not only suffers from significant class imbalance but also presents substantial geometric similarity between certain classes (Fig.~\ref{fig:framework} (left))}.  
Specifically, the number of samples per class varies considerably: the class \textit{chair} contains $\sim 900$ samples, whereas the class \textit{bowl} has over $10 \times$ fewer.  
Moreover, as highlighted in our case study in Sec. \ref{sec:case_study}, some classes exhibit strong geometric resemblance, such as \textit{night stand} and \textit{dresser}, or \textit{desk} and \textit{table}, \textit{vase} and \textit{flower pot}.  

These intrinsic imbalances and similarities hinder models from learning well-disentangled features, as dominant classes may bias feature learning, while geometrically similar classes tend to cluster in feature space.  
The fixed ETF head exacerbates these issues, as its global orientation is fixed once initialized, preventing adaptation to the underlying data distribution. 
To address these challenges, we propose an adaptive training framework.

\subsubsection{Representation-Balanced Learning (RBL)}
The classification matrix $\mathbf{W}$ in a fixed ETF head is untrainable. 
Its orientation is dictated by the rotation matrix $\mathbf{R}$ in Eq. \ref{eq:etf}.
However, the orientation can critically influence a model's generalization performance \cite{peifeng2023feature}. 
Our ablation study (Sec. \ref{sec:ablation}) shows that using a fixed ETF head harms classification accuracy on benign samples.

To address this issue, RBL enables the ETF head to be learnable, albeit under specific constraints. 
RBL was initially introduced to enhance the clean performance of models employing a fixed ETF head in long-tailed scenarios \cite{peifeng2023feature}. 
In this approach, the rotation matrix $\mathbf{R}$ is updated during training, but is constrained to remain an orthogonal matrix, preserving the ETF properties (Fig. \ref{fig:framework} (b)). 
This constraint can be enforced with standard optimization algorithms such as SGD or Adam.
Consequently, both the feature extractor and the classifier jointly optimize feature representations and global orientation, reaching balanced clean performance on imbalanced point cloud datasets.
RBL also forms more accurate feature representations, which is crucial for FDL below.

\subsubsection{Dynamic Feature Direction Loss (FDL)}
While RBL enhances feature alignment, it does not fully address the challenges in complex point cloud datasets. 
In particular, point clouds often share overlapping geometric features between classes, making it challenging for ETF and RBL to achieve adequate feature disentanglement. 

To better handle these characteristics, we introduce the Feature Direction Loss (FDL), which dynamically adjusts the direction of sample features during training. 
FDL encourages each sample to align more closely with its within-class mean $\mathbf{\bar{h}}_k$. 
Additionally, it promotes repulsion by pushing sample features away from the mean of the closest non-ground-truth class $\mathbf{\bar{h}}_{k'}$, where $k' \neq k$ (Fig. \ref{fig:framework} (c)).

Formally, the dynamic FDL is defined as:
\begin{equation} \label{eq:fdl_loss}
    \mathcal{L}_{FDL}(\mathbf{h}, \bar{\mathbf{h}}_k, \bar{\mathbf{h}}_{k'}) = -\frac{\mathbf{h}^T \bar{\mathbf{h}}_k}{\|\mathbf{h}\| \|\bar{\mathbf{h}}_k\|} + \frac{\mathbf{h}^T \bar{\mathbf{h}}_{k'}}{\|\mathbf{h}\| \|\bar{\mathbf{h}}_{k'}\|},
\end{equation}
where $\|\cdot\|$ denotes the $L_2$ norm, $\bar{\mathbf{h}}_k, \bar{\mathbf{h}}_{k'}$ are updated at every epoch.
This formulation clusters intra-class features while enhancing inter-class repulsion by reducing similarity with non-ground-truth class means.

\subsection{Overall Framework}
The overall framework of our proposed {\projname} is illustrated in Fig.~\ref{fig:framework} (right). 
The training objective is defined as:
\begin{equation}
    \mathcal{L} = \mathcal{L}_{dot}(\mathbf{h}, \mathbf{W}) + \lambda \cdot \mathcal{L}_{FDL}(\mathbf{h}, \bar{\mathbf{h}}_k, \bar{\mathbf{h}}_{k'}),
\end{equation}
where $\lambda$ is a hyperparameter that controls the weight of $\mathcal{L}_{FDL}$. 
To ensure that $\mathcal{L}_{FDL}$ provides accurate guidance, it is only incorporated after several warm-up epochs. 
Detailed pseudocode is provided in Appendix \ref{sec:algorithm}.

During training, sample features undergo global rotation by RBL, while FDL further aligns them with their respective class means. 
RBL dynamically provides corrected orientations for ETF, where the adapted sample features is used by FDL to refine the feature space.
This dual adjustment enhances the model’s capability to distinguish between classes, even in scenarios of high inter-class geometric similarity.
Combining the strengths of the ETF head with the adaptive training framework, we construct a robust model that naturally promotes well-disentangled feature extraction, addressing the unique characteristics of point cloud data.

Importantly, {\projname} is model-agnostic, which can be integrated with any point cloud recognition model. 
By simply replacing the classifier with a simplex ETF head and training the model using the adaptive training framework, robust models can be developed with minimal computational overhead. 
Additionally, {\projname} can be combined with input preprocessing defenses to further enhance robustness.

\begin{table*}[t]
    \centering

    \setlength{\tabcolsep}{2pt}
    \resizebox{0.95\linewidth}{!}{
        
    \begin{tabular}{c@{\hspace{8pt}} cccccc@{\hspace{8pt}} cccccc@{\hspace{8pt}} cccccc}
    \toprule
	\multicolumn{1}{c}{\multirow{2}{*}{Defense}} & \multicolumn{6}{c}{PointNet}		& \multicolumn{6}{c}{DGCNN}		& \multicolumn{6}{c}{PCT}\\
	\cmidrule(r{8pt}){2-7} \cmidrule(r{8pt}){8-13} \cmidrule(r){14-19}
    & IFGM 	& G3-Pert 	& G3-Add 	& Drop 	& AdvPC 	& Avg. 	& IFGM 	& G3-Pert 	& G3-Add 	& Drop 	& AdvPC 	& Avg. 	& IFGM 	& G3-Pert 	& G3-Add 	& Drop 	& AdvPC 	& Avg. \\ 
	\midrule
    \rowcolor{TableYellow!25}
	Vanilla    	 & 1.6	 & 64.4	 & 48.7	 & 65.0	 & 17.6	 & 39.5	 & 0.0	 & 34.6	 & 27.4	 & 69.6	 & 4.3	 & 27.2	 & 7.4	 & 64.7	 & 56.5	 & 82.9	 & 26.2	 & 47.5\\
	\rowcolor{TableGreen!20} SOR$^{\dag}$        	 & 23.6	 & 63.4	 & 50.9	 & 69.0	 & 20.9	 & 45.6	 & 11.9	 & 35.2	 & 25.2	 & 71.6	 & 13.2	 & 31.4	 & 38.3	 & 63.7	 & 54.9	 & 79.0	 & 32.2	 & 53.6\\
	\rowcolor{TableGreen!20} DUP-Net$^{\dag}$    	 & 24.8	 & 63.0	 & 50.4	 & 67.8	 & 22.0	 & 45.6	 & 32.1	 & 37.1	 & 20.7	 & 49.2	 & 13.7	 & 30.5	 & 45.1	 & 62.5	 & 52.2	 & 73.9	 & 36.9	 & 54.1\\
	\rowcolor{TableGreen!20} Diffusion$^{\dag}$  	 & 33.2	 & 64.4	 & 49.2	 & 69.1	 & 23.5	 & 47.9	 & 63.4	 & 39.8	 & 29.5	 & 68.6	 & 33.2	 & 46.9	 & 62.2	 & 64.7	 & 57.9	 & \textbf{84.1}	 & 38.4	 & 61.5\\
	\rowcolor{TableBlue1!10} AT$^{\circ}$         	 & 15.1	 & 48.1	 & 22.0	 & 36.3	 & 4.9	 & 25.3	 & 22.7	 & 28.6	 & 30.3	 & 59.1	 & 2.5	 & 28.6	 & 5.1	 & 59.2	 & 57.4	 & 79.8	 & 5.3	 & 41.4\\
	\rowcolor{TableBlue1!10} PAGN$^{\circ}$       	 & 24.1	 & 63.5	 & 50.3	 & 69.5	 & 19.5	 & 45.4	 & 10.5	 & 34.9	 & 21.5	 & 71.2	 & 6.1	 & 28.8	 & 49.5	 & 52.1	 & 52.5	 & 75.1	 & 33.9	 & 52.6\\
	\rowcolor{TableBlue1!10} PointCutMix$^{\circ}$	 & 32.4	 & 45.8	 & 44.2	 & 73.6	 & 9.8	 & 41.2	 & 37.4	 & 43.6	 & 43.2	 & 63.3	 & 19.5	 & 41.4	 & 58.3	 & 41.9	 & 47.8	 & 74.2	 & 23.1	 & 49.0\\
	\rowcolor{TableBlue1!10} CAP$^{\circ}$        	 & 25.5	 & 65.4	 & 49.9	 & 59.2	 & 16.3	 & 43.3	 & 2.0	 & 58.5	 & 36.4	 & 78.7	 & 15.4	 & 38.2	 & 45.7	 & 47.7	 & 43.4	 & 76.0	 & 13.8	 & 45.3\\
	\rowcolor{TableBlue1!25} Ours$^{\circ}$       	 & \textbf{80.2}	 & \textbf{79.8}	 & \textbf{75.6}	 & \textbf{80.4}	 & \textbf{77.9}	 & \textbf{78.8}	 & \textbf{84.3}	 & \textbf{84.2}	 & \textbf{69.9}	 & \textbf{84.6}	 & \textbf{81.3}	 & \textbf{80.9}	 & \textbf{75.2}	 & \textbf{81.1}	 & \textbf{72.6}	 & 82.2	 & \textbf{75.5}	 & \textbf{77.3}\\
    \bottomrule
    \end{tabular}
    }

    \caption{Classification accuracy (\%) with defenses (\colorbox{TableGreen!20}{input preprocessing$^{\dag}$
    }, \colorbox{TableBlue1!10}{self-robust$^{\circ}$}) on ModelNet40. Best values in bold. }
    
    \label{table:main_mn40}
\end{table*}

\begin{table}[t]

    \setlength{\tabcolsep}{2pt}
    \resizebox{\linewidth}{!}{
    \begin{tabular}{c >{\columncolor{TableYellow!25}}c >{\columncolor{TableGreen!20}}c>{\columncolor{TableGreen!20}}c>{\columncolor{TableGreen!20}}c >{\columncolor{TableBlue1!10}}c>{\columncolor{TableBlue1!10}}c>{\columncolor{TableBlue1!10}}c>{\columncolor{TableBlue1!10}}c >{\columncolor{TableBlue1!25}}c}
    \toprule
	Attack  & Vanilla	& SOR$^{\dag}$	& DUP-Net$^{\dag}$	& Diffusion$^{\dag}$	& AT$^{\circ}$	& PAGN$^{\circ}$	& PointCutMix$^{\circ}$	& CAP$^{\circ}$	& Ours\\ 
	\midrule
	KNN        	 & 30.2	 & 36.6	 & 37.2	 & \underline{50.0}	 & 4.8	 & 36.1	 & 27.8	 & 32.0	 & \textbf{79.9}\\
	GeoA3      	 & 8.1	 & 51.3	 & 51.0	 & \textbf{72.2}	 & 1.3	 & 49.2	 & 54.7	 & 49.3	 & \underline{55.3}\\
	SI	 & 18.7	 & 35.5	 & 34.1	 & \underline{45.6}	 & 4.6	 & 40.2	 & 44.2	 & 29.4	 & \textbf{53.0}\\
	HiT        	 & 41.9	 & 51.4	 & 52.8	 & 51.9	 & 35.8	 & 48.5	 & \textbf{73.7}	 & 51.5	 & \underline{66.1}\\
	Avg.       	 & 24.7	 & 43.7	 & 43.8	 & \underline{54.9}	 & 11.6	 & 43.5	 & 50.1	 & 40.6	 & \textbf{63.6}\\
\bottomrule
    \end{tabular}
    }
    \caption{Classification accuracy (\%) of shape-invariant attacks against defenses on ModelNet40, PointNet. Best values in bold, second-best values underlined.}
    \label{table:untarget_shape_invar_PointNet}
\end{table}

\subsection{Robustness Analysis} \label{sec:robust_analysis}
We further analyze how {\projname} produces a robust feature extractor characterized by cohesive intra-class features and distinct inter-class features. 
The analysis is based on the gradient of loss functions with respect to sample feature $\mathbf{h}$.

We start with $\mathcal{L}_{dot}$, whose gradient is given by:
\begin{equation}
    \nabla_{\mathbf{h}} \mathcal{L}_{dot}(\mathbf{h}, \mathbf{W}) = \left(\cos \angle(\mathbf{h}, \mathbf{w}_k) - 1\right) \mathbf{w}_k.
\end{equation}
Thus, $\mathcal{L}_{dot}$ pulls the feature $\mathbf{h}$ to align with the predefined classification vector $\mathbf{w}_k$ in the ETF head. 
According to NC3, $\mathbf{w}_k$ shares the same direction as the mean feature of class $k$.

Next, we analyze the loss function $\mathcal{L}_{FDL}$:
\begin{align}
    \nabla_{\mathbf{h}} \mathcal{L}_{FDL} &= \nabla_{\text{pull}} + \nabla_{\text{push}}, \\
    \nabla_{\text{pull}} &= \frac{(\mathbf{h}^T \bar{\mathbf{h}}_k) \mathbf{h}}{\|\mathbf{h}\|^3 \|\bar{\mathbf{h}}_k\|} - \frac{\bar{\mathbf{h}}_k}{\|\mathbf{h}\| \|\bar{\mathbf{h}}_k\|}, \\
    \nabla_{\text{push}} &= \frac{\bar{\mathbf{h}}_{k'}}{\|\mathbf{h}\| \|\bar{\mathbf{h}}_{k'}\|} - \frac{(\mathbf{h}^T \bar{\mathbf{h}}_{k'}) \mathbf{h}}{\|\mathbf{h}\|^3 \|\bar{\mathbf{h}}_{k'}\|}.
\end{align}    
Hence, $\mathcal{L}_{FDL}$ promotes the alignment of feature $\mathbf{h}$ with the class $k$ mean, while simultaneously pushing it away from the closest non-ground-truth class $k'$ feature mean. 
The degree of repulsion is adaptively scaled by the similarity between the two classes, thereby strengthening the separation of challenging class pairs.

Since $\bar{\mathbf{h}}_k$ and $\bar{\mathbf{h}}_{k'}$ are derived from actual sample features extracted by model, $\mathcal{L}_{FDL}$ provides a practical complement to $\mathcal{L}_{dot}$: \emph{it balances the ideal simplex ETF structure with a more flexible yet well-disentangled feature organization}. 
Notably, the effectiveness of this complement depends on accurate $\bar{\mathbf{h}}_k$ and $\bar{\mathbf{h}}_{k'}$. 
To address potential biases in feature distribution, the global rotation by RBL mitigates imbalances, indirectly enhancing feature disentanglement.  
In summary, adversarial robustness is initially enhanced by the ETF classifier and is further reinforced by $\mathcal{L}_{FDL}$ with RBL, which explicitly separates features that are closely clustered. 
Further details are provided in Appendix \ref{sec:supple_robust_analysis}.

\section{Experiments} \label{sec:exp}
In this section, we address the following research questions:

\noindent
\textbf{RQ1}: How robust is {\projname} against both standard adversarial attacks and shape-invariant attacks?

\noindent
\textbf{RQ2}: How do the ETF head and the adaptive training framework (RBL, FDL) contribute to {\projname}?

\noindent
\textbf{RQ3}: How does feature disentanglement correlate with robustness? To what extent does {\projname} improve the feature disentanglement capabilities of existing models?

\subsection{Experimental Settings}

\textbf{Models and Datasets.}
We evaluate three point cloud classification models with different architectures: PointNet \cite{qi2017pointnet}, DGCNN \cite{wang2019dynamic}, Point Cloud Transformer (PCT) \cite{guo2021pct} using ModelNet40 \cite{wu20153dshapenet} (with $9,843$ training, $2,468$ test samples) and ShapeNet \cite{chang2015shapenet} (with $35,708$ training, $15,429$ test samples). Each object is sampled to $2,048$ points and normalized to a unit sphere. For efficiency, we randomly select $2,732$ samples from ShapeNet's test set, with a maximum of $60$ samples per class. 

\noindent\textbf{Baseline Attacks.}
We test nine attacks: IFGM \cite{liu2019extending}, G3-Add \cite{xiang2019generating}, G3-Pert \cite{xiang2019generating}, Drop \cite{zheng2019pointcloudsaliency}, AdvPC \cite{hamdi2020advpc}, KNN \cite{tsai2020robustknn}, GeoA3 \cite{wen2020geometry}, ShapeInvariant (SI) \cite{huang2022shape}, and HiT \cite{lou2024hide}. 
We consider untargeted attacks, i.e., the attacker aims at suppressing the prediction of the ground truth label. 

\noindent\textbf{Baseline Defenses.}
We include three input-preprocessing defenses: SOR \cite{zhou2019dupnet}, DUP-Net \cite{zhou2019dupnet}, and PointDP (Diffusion) \cite{sun2023critical}, and four self-robust models: AT \cite{liu2019extending}, PAGN \cite{liang2022pagn}, PointCutMix(K) \cite{zhang2022pointcutmixa}, and CAP \cite{ding2023cap}. 

\noindent\textbf{Evaluation Metrics.}
We use classification accuracy (ACC) as evaluation metric of robustness. We use Silhouette Coefficient (SC) to quantify the quality of the model feature space. More details 
are available in Appendix \ref{sec:supple_exp_settings}.

\subsection{Robustness against Attacks}
To address \textbf{RQ1}, we first evaluate the robustness of {\projname} against various attacks.
The results for ModelNet40 are presented in Table \ref{table:main_mn40}, while the results for ShapeNet are provided in Appendix \ref{sec:more_results}. 
The \textit{Vanilla} row represents models with no defense. The \textit{Avg.} columns summarize the average results for each row.

Table \ref{table:main_mn40} shows that \textit{existing defenses struggle with unseen attacks.} 
Among input preprocessing defenses, DUP-Net performs poorly against Drop attacks, as its reconstruction capabilities cannot restore missing parts of a point cloud, resulting in lowest ACC on both DGCNN and PCT. 
For self-robust models, AT fails to generalize to unseen attacks like AdvPC, leading to an ACC of $2.5\%$ on DGCNN. 
Similar performance drops share with PointCutMix and CAP. 
In contrast, \textit{{\projname} excels due to the inherently well-disentangled feature modeling provided by NC.} It achieves state-of-the-art performance against most attacks across three models. For example, the average ACC of DGCNN increases substantially from $27.2\%$ to $80.9\%$, surpassing second-best baseline by $34.0\%$.

Additionally, we evaluate shape-invariant attacks on PointNet in Table \ref{table:untarget_shape_invar_PointNet}. 
\textit{The results echo that existing defenses struggle with unseen adversarial patterns.}
For instance, Diffusion performs well against attacks with subtle perturbations along the object surface, such as GeoA3, yet fails to counteract the hidden noise beneath the surface introduced by HiT. 
Conversely, while PointCutMix is robust against HiT, its self-supervised learning cannot survive more severe perturbations. 
On the contrary, \textit{{\projname} achieves best overall robustness by enforcing the robustly disentangled feature space.}
Results for DGCNN, PCT are in Appendix \ref{sec:more_defense_tables}.

\begin{table}[t]

    \footnotesize \resizebox{\linewidth}{!}{
    
    \begin{tabular}{c c ccccc c}
        \toprule   Component      & Clean            & IFGM              & G3-Pert        & G3-Add         & Drop              & AdvPC             & Avg.              \\
        \midrule Vanilla & 86.2             & 1.6               & 64.4              & 48.7              & 65.0              & 17.6              & 39.5              \\
        + ETF            & -0.6             & \underline{+76.9} & +13.9             & \underline{+25.0} & \underline{+14.2} & \textbf{+60.6}    & \underline{+38.1} \\
        + ETF \& RBL     & \textbf{+3.7}    & +66.7             & \underline{+14.4} & +24.4             & +7.9              & +59.2             & +34.5             \\
        + ETF \& FDL     & -1.4             & +66.5             & +9.9              & +22.7             & +4.9              & +54.3             & +31.7             \\
        + All (Ours)     & \underline{+0.9} & \textbf{+78.5}    & \textbf{+15.4}    & \textbf{+27.0}    & \textbf{+15.4}    & \underline{+60.3} & \textbf{+39.3}    \\
        \bottomrule
    \end{tabular}
    }

    \caption{Classification accuracy (\%) of attacks against {\projname}
    with different components on ModelNet40, PointNet. Best values in bold, second-best values underlined. }
    \label{table:ablation} 
\end{table}

\subsection{Ablation Study} \label{sec:ablation}

To address \textbf{RQ2}, we systematically investigate the impact of each component in {\projname} using PointNet as follows.

\noindent
\textbf{ETF}  
(1) The ETF classifier \textit{significantly enhances robustness} by enforcing feature orthogonality. The average ACC of \underline{ETF} improves by $38.1\%$ compared to \underline{Vanilla}.  
(2) However, the fixed global orientation limits adaptability on imbalanced data, resulting in a $0.6\%$ drop in clean ACC, suggesting \textit{a trade-off between robustness and generalization}.

\noindent
\textbf{RBL}  
(1) RBL facilitates ETF by allowing learnable rotation, enabling ETF to better align with the sample features and thereby \textit{preventing clean performance degradation}, as seen from the $3.7\%$ improvement from \underline{ETF} to \underline{ETF \& RBL}.  
(2) However, it also introduces \textit{instability in feature consistency}, which weakens adversarial robustness. 

\noindent
\textbf{FDL}  
FDL promotes intra-class clustering and inter-class repulsion, yet \textit{its effectiveness depends on the feature quality}.  
(1) \underline{ETF} vs. \underline{ETF \& FDL}: We observe that with misaligned features, FDL leads to suboptimal feature directions, reducing both clean ACC and robustness.  
(2) \underline{ETF \& RBL} vs. \underline{All}: Conversely, when RBL provides better-aligned features, FDL effectively enhances class separation, thereby improving robustness. 
(3) With more advanced architectures like DGCNN and PCT, where feature space is more structured, FDL further improves both clean ACC and robustness. Results and visualizations can be found in Appendix \ref{sec:more_ablations}, \ref{sec:more_defense_visual}.

\subsection{Quality of Feature Space}  \label{sec:feat_qlty}

To address \textbf{RQ3}, we analyze the correlation between feature space quality and adversarial robustness, as depicted in Fig. \ref{fig:feat_qlty}.  
Specifically, we quantify feature quality using \textit{Silhouette Coefficient (SC)}, which captures intra-class compactness and inter-class separability. Robustness is measured by average ACC in Table \ref{table:main_mn40}.  
The results indicate that a higher-quality feature space substantially enhances robustness, as adversarial perturbations struggle to shift samples across well-separated decision boundaries.
Our {\projname} can effectively improve the feature quality of existing models.

\subsection{More Results}

\begin{figure}[t]
    \centering
    \includegraphics[width=0.8\linewidth]{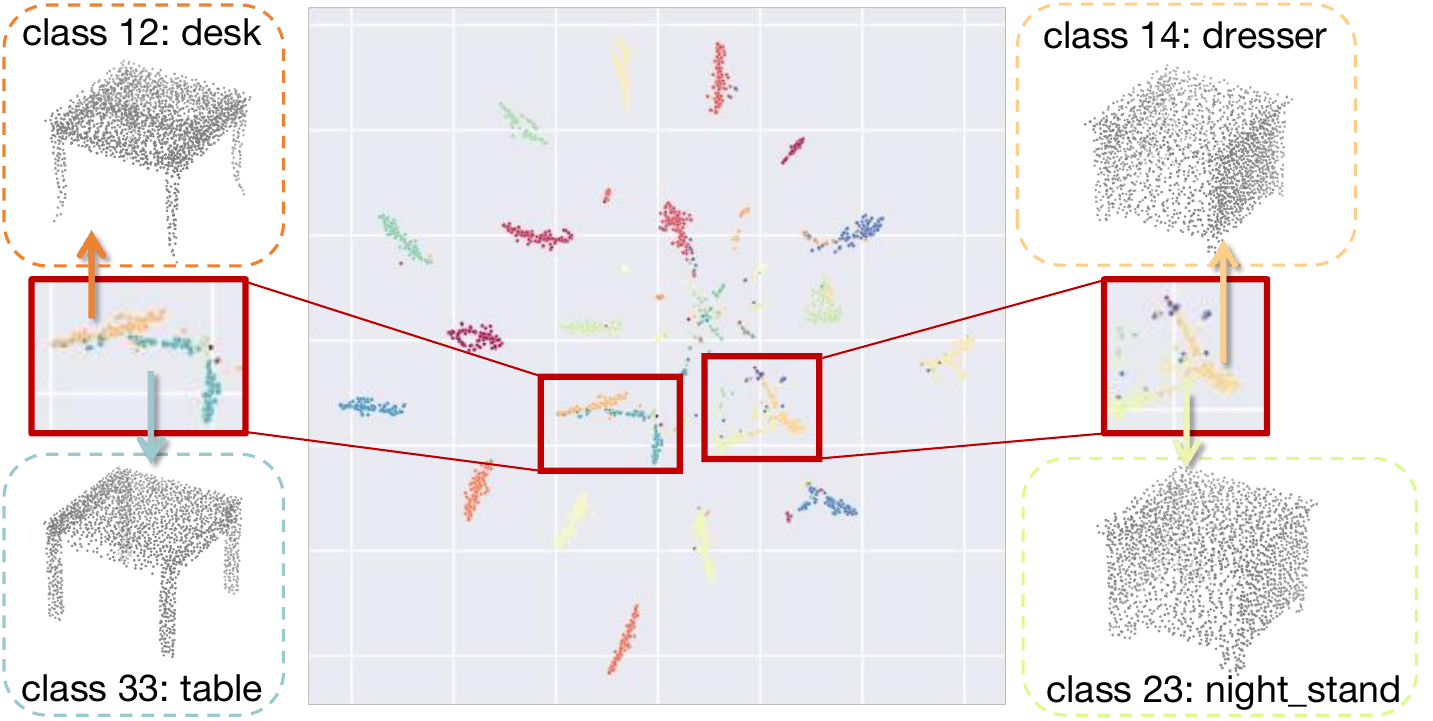}
    \caption{Case study for highly similar classes.}
    \label{fig:case_study}
\end{figure}

\noindent
\textbf{Case Study}  \label{sec:case_study}
Fig. \ref{fig:model_tsne} illustrates that the feature distributions of certain classes frequently overlap, as exemplified by the orange and cyan classes in the lower center of the \underline{All} subplot. 
We further show how inter-class geometric similarity can challenge the training of a raw NC model.

Specifically, in Fig. \ref{fig:case_study}, we examine two pairs of classes: \textit{desk} vs. \textit{table} and \textit{night stand} vs. \textit{dresser}. 
Given the high similarity between samples in these class pairs, the feature extractor naturally encounters difficulties in distinguishing them, a task that even humans might struggle with.

From Fig. \ref{fig:model_tsne}, {\projname} makes significant efforts to help the feature extractor differentiate between these classes, leading to partially separated feature strips. 
However, \textit{such inherent dataset flaws limit defenses from achieving optimal robustness with ideal feature separation}.

\noindent
\textbf{Efficiency}
{\projname} replaces the classifier with an ETF head, which introduces almost no additional cost compared to other defenses, as detailed in Appendix \ref{sec:efficiency}.

\section{Conclusion}
In summary, our work introduces Neural Collapse for robust point cloud recognition. Our analysis reveals that existing models and defenses suffer from inadequate feature disentanglement, which renders them susceptible to adversarial attacks. 
To this end, we propose {\projname}, which replaces model classifier with an ETF head to achieve feature disentanglement.
Recognizing the unique challenges posed by class imbalance and complex geometric similarities in point cloud data, we further propose an adaptive training framework that integrates RBL and dynamic FDL. 
The framework effectively boosts robustness, surpassing current defenses.
Moreover, {\projname} is model-agnostic and efficient, allowing for integration with any existing model architecture.

\clearpage
\setcounter{secnumdepth}{0}
\renewcommand\thesubsection{\arabic{subsection}}
\renewcommand\labelenumi{\thesubsection.\arabic{enumi}}

\section{Acknowledgements}
We would like to thank the anonymous reviewers for their insightful comments that helped improve the quality of the paper.
We also appreciate the researchers who contributed to the collection and curation of public datasets.
This work was supported in part by the National Natural Science Foundation of China (62472096, 62172104, 62172105, 62102093, 62102091, 62302101, 62202106, 6250074640).
Min Yang is a faculty of the Shanghai Institute of Intelligent Electronics \& Systems and Engineering Research Center of Cyber Security Auditing and Monitoring, Ministry of Education, China.

\setcounter{secnumdepth}{2}
\renewcommand\thesubsection{\thesection.\arabic{subsection}}
\renewcommand\labelenumi{\theenumi}

\bibliography{aaai2026}

\setcounter{secnumdepth}{2}
\clearpage
\appendix
\section{Details of Robustness Analysis} \label{sec:supple_robust_analysis}

In this section, we provide an in-depth exploration of the robustness analysis as presented in Sec. \ref{sec:robust_analysis} of the paper. 
Specifically, we focus on the gradient analysis of both the dot loss and the dynamic feature direction loss utilized in the robustness evaluation.

\subsection{The Dot Loss}

We begin by revisiting the definition of the dot loss:
\begin{equation} \label{eq:dot_loss_apdx}
    \mathcal{L}_{dot}(\mathbf{h}, \mathbf{W}) = \frac{1}{2\sqrt{E_W E_H}} \left(\mathbf{w}_k^T \mathbf{h} - \sqrt{E_W E_H}\right)^2,
\end{equation}
where \(\mathbf{h}\) denotes the feature vector of a sample from class \(k\), and \(\mathbf{w}_k\) represents the corresponding classification vector. The constants \(E_W\) and \(E_H\) are predefined \(L_2\)-norm constraints, ensuring that \(||\mathbf{h}||^2 \le E_H\) and \(||\mathbf{w}_k||^2 \le E_W\).

To understand how the dot loss influences the feature extractor during training, we compute the partial derivative of \(\mathcal{L}_{dot}\) with respect to the feature \(\mathbf{h}\):
\begin{align}
    \nabla_{\mathbf{h}} \mathcal{L}_{dot} &= \nabla_{\mathbf{h}} \left[ \frac{1}{2\sqrt{E_W E_H}} \left(\mathbf{w}_k^T \mathbf{h} - \sqrt{E_W E_H}\right)^2 \right] \nonumber \\
    &= \frac{1}{\sqrt{E_W E_H}} \left(\mathbf{w}_k^T \mathbf{h} - \sqrt{E_W E_H}\right) \mathbf{w}_k.
\end{align}

Next, we express the cosine similarity between \(\mathbf{h}\) and \(\mathbf{w}_k\) as:
\[
    \cos \angle(\mathbf{h}, \mathbf{w}_k) = \frac{\mathbf{w}_k^T \mathbf{h}}{\|\mathbf{w}_k\| \|\mathbf{h}\|} = \frac{\mathbf{w}_k^T \mathbf{h}}{\sqrt{E_W E_H}}.
\]
Substituting this expression into the gradient equation, we obtain:
\begin{equation}
    \nabla_{\mathbf{h}} \mathcal{L}_{dot} = \left(\cos \angle(\mathbf{h}, \mathbf{w}_k) - 1\right) \mathbf{w}_k.
\end{equation}

\subsection{The Dynamic Feature Direction Loss}

The dynamic feature direction loss is defined as:
\begin{equation} \label{eq:fdl_loss_apdx}
    \mathcal{L}_{FDL}(\mathbf{h}, \bar{\mathbf{h}}_k, \bar{\mathbf{h}}_{k'}) = -\frac{\mathbf{h}^T \bar{\mathbf{h}}_k}{\|\mathbf{h}\| \|\bar{\mathbf{h}}_k\|} + \frac{\mathbf{h}^T \bar{\mathbf{h}}_{k'}}{\|\mathbf{h}\| \|\bar{\mathbf{h}}_{k'}\|},
\end{equation}
where $\|\cdot\|$ denotes the $L_2$ norm, and $\bar{\mathbf{h}}_k$ and $\bar{\mathbf{h}}_{k'}$ represent the mean feature vectors of the ground-truth class $k$ and the nearest non-ground-truth class $k'$, respectively. $\bar{\mathbf{h}}_k$ and $\bar{\mathbf{h}}_{k'}$ are updated at every epoch.

To understand the influence of $\mathcal{L}_{FDL}$ on the feature $\mathbf{h}$ during training, we compute its partial derivative with respect to $\mathbf{h}$ as follows.

For the first term \(-\frac{\mathbf{h}^T \bar{\mathbf{h}}_k}{\|\mathbf{h}\| \|\bar{\mathbf{h}}_k\|}\), the gradient is given by:
\begin{align}
\nabla_{\mathbf{h}} \left(-\frac{\mathbf{h}^T \bar{\mathbf{h}}_k}{\|\mathbf{h}\| \|\bar{\mathbf{h}}_k\|}\right) &= -\frac{1}{\|\bar{\mathbf{h}}_k\|} \left(\frac{\bar{\mathbf{h}}_k \|\mathbf{h}\| - \frac{\mathbf{h}^T \bar{\mathbf{h}}_k}{\|\mathbf{h}\|} \mathbf{h}}{\|\mathbf{h}\|^2}\right) \nonumber \\
&= -\frac{\bar{\mathbf{h}}_k}{\|\mathbf{h}\| \|\bar{\mathbf{h}}_k\|} + \frac{\mathbf{h}^T \bar{\mathbf{h}}_k}{\|\mathbf{h}\|^3 \|\bar{\mathbf{h}}_k\|} \mathbf{h}.
\end{align}

Similarly, for the second term \(\frac{\mathbf{h}^T \bar{\mathbf{h}}_{k'}}{\|\mathbf{h}\| \|\bar{\mathbf{h}}_{k'}\|}\), the gradient is:
\begin{align}
\nabla_{\mathbf{h}} \left(\frac{\mathbf{h}^T \bar{\mathbf{h}}_{k'}}{\|\mathbf{h}\| \|\bar{\mathbf{h}}_{k'}\|}\right) &= \frac{1}{\|\bar{\mathbf{h}}_{k'}\|} \left(\frac{\bar{\mathbf{h}}_{k'} \|\mathbf{h}\| - \frac{\mathbf{h}^T \bar{\mathbf{h}}_{k'}}{\|\mathbf{h}\|} \mathbf{h}}{\|\mathbf{h}\|^2}\right) \nonumber \\
&= \frac{\bar{\mathbf{h}}_{k'}}{\|\mathbf{h}\| \|\bar{\mathbf{h}}_{k'}\|} - \frac{\mathbf{h}^T \bar{\mathbf{h}}_{k'}}{\|\mathbf{h}\|^3 \|\bar{\mathbf{h}}_{k'}\|} \mathbf{h}.
\end{align}

Combining these gradients, we obtain the total gradient of $\mathcal{L}_{FDL}$ as:
\begin{align}
    \nabla_{\mathbf{h}} \mathcal{L}_{FDL} &= \nabla_{\text{pull}} + \nabla_{\text{push}}, \\
    \nabla_{\text{pull}} &= \frac{(\mathbf{h}^T \bar{\mathbf{h}}_k) \mathbf{h}}{\|\mathbf{h}\|^3 \|\bar{\mathbf{h}}_k\|} - \frac{\bar{\mathbf{h}}_k}{\|\mathbf{h}\| \|\bar{\mathbf{h}}_k\|}, \\
    \nabla_{\text{push}} &= \frac{\bar{\mathbf{h}}_{k'}}{\|\mathbf{h}\| \|\bar{\mathbf{h}}_{k'}\|} - \frac{(\mathbf{h}^T \bar{\mathbf{h}}_{k'}) \mathbf{h}}{\|\mathbf{h}\|^3 \|\bar{\mathbf{h}}_{k'}\|}.
\end{align}

\section{Implementation Details}
\subsection{The Proposed {\projname}} \label{sec:algorithm}
We further present the implementation details of our proposed {\projname}. 
The overall training process is summarized in Algorithm \ref{alg:model}.

\renewcommand{\algorithmicrequire}{\textbf{Input:}}
\renewcommand{\algorithmicensure}{\textbf{Output:}}

\begin{algorithm}[h]
    \caption{Train a model with the proposed {\projname}. }
    \label{alg:model}
    \begin{algorithmic}[1]
    \Require{Initial model $F=f \circ g$, train set $(\mathbf{X}, y) \in \mathcal{D}_\text{Train}$, learning rate $\eta$, total train epochs $E$, start epoch $E_{FDL}$ of $\mathcal{L}_{FDL}$, weight $\lambda$ of $\mathcal{L}_{FDL}$. }
    \Ensure{Robust model $F$. }

    \State Initialize learnable ETF head $g(\mathbf{h}) = \mathbf{h}\mathbf{W}$.   \Comment{Def.1. }
    \State Regulate rotation $\mathbf{R}$ in $\mathbf{W}$ to be orthogonal.      

    \For{i in range($1, E_{FDL}$)}     \Comment{Training with $\mathcal{L}_{dot}$ only. }
        \For{$(\mathbf{X}, y) \in \mathcal{D}_\text{Train}$}    
        \State $\mathbf{h} \gets f(\mathbf{X})$.
        \State $\ell_{dot} \gets \mathcal{L}_{dot}(\mathbf{h},\mathbf{W})$.
        \State Update $f, g$ with $\ell_{dot}, \eta$.
        \EndFor
    \EndFor

    \For{i in range($E_{FDL}, E$)}     \Comment{Add $\mathcal{L}_{FDL}$ to training. }
        \State Compute class feature centroids $\bar{\mathbf{h}}_k$ on $\mathcal{D}_\text{Train}$.
        \For{$(\mathbf{X}, y) \in \mathcal{D}_\text{Train}$}    
        \State $\mathbf{h} \gets f(\mathbf{X})$.
        \State $\ell_{dot} \gets \mathcal{L}_{dot}(\mathbf{h},\mathbf{W})$.
        \State $\ell_{FDL} \gets \mathcal{L}_{FDL}(\mathbf{h}, \bar{\mathbf{h}}_k, \bar{\mathbf{h}}_{k'})$.
        \State $\ell = \ell_{dot} + \lambda \cdot \ell_{FDL}$.
        \State Update $f, g$ with $\ell, \eta$.
        \EndFor
    \EndFor
    
    \Return Robust model with ETF head $F=f\circ g$. 
\end{algorithmic}
\end{algorithm}

The training of a {\projname} model begins with the initialization of an ETF classification head, which follows Definition 1 in the main paper. 
To allow the rotation of the ETF head under constraints during training, we use the geotorch library to regulate matrix $\mathbf{R}$ in $\mathbf{W}$ following \cite{peifeng2023feature}. 
The training process involves two stages: the warm-up stage with $\mathcal{L}_{dot}$ only, and the enhancing stage with both $\mathcal{L}_{dot}$ and $\mathcal{L}_{FDL}$. 
To mitigate the impact of extreme outliers within an observed point cloud during inference, we also incorporate the SOR algorithm as a pre-processing step before classification. 

For hyperparameters, we train the model for $E=60$ epochs in total, with a $E_{FDL}=10$ epochs of warm-up. We set learning rate $\eta=0.001$, weight $\lambda=5$.
We use the default settings of $\alpha=1.1$ and top-$k=2$ for the SOR algorithm \cite{wang2019dynamic}, as proposed in the original work. 
For other hyperparameters of existing point cloud classification models \cite{qi2017pointnet,wang2019dynamic,li2018pointcnn}, we follow the specifications provided in the respective original works.

\subsection{Experimental Settings} \label{sec:supple_exp_settings}
We further detail the background and the hyper-parameter settings of the baseline attack and defense strategies we used in our paper. 

\noindent\textbf{Attack Methods. }
We briefly summarize the attack methods involved in our work. 
Nine attacks are considered in this paper, including IFGM \cite{liu2019extending}, G3-Add \cite{xiang2019generating}, G3-Pert \cite{xiang2019generating}, Drop \cite{zheng2019pointcloudsaliency}, AdvPC \cite{hamdi2020advpc}, KNN \cite{tsai2020robustknn}, GeoA3 \cite{wen2020geometry}, ShapeInvariant (SI) \cite{huang2022shape}, and HiT \cite{lou2024hide}. 
Specifically, IFGM exploits normalized gradients to generate perturbations instead of the product of the sign of gradients and a perturbation clip threshold originally used in FGSM \cite{goodfellow2014explaining}. 
G3-Pert and G3-Add stand for the perturbation and adding attacks proposed in the original work, where G3-Pert imposes small perturbations on all points of a point cloud under the constraint of $L_2$ distance, while G3-Add imposes the perturbations on a copy of a separate subset of points from the input point cloud, considered as the added points, under the constraint of Chamfer distance. 
Drop attack selects the most important set of points that influences the final extracted feature for a point cloud and deletes them. 
AdvPC generates an adversarial point cloud given the benign one with an autoencoder trained for a dataset with the $L_2$ distance constraint. 
KNN is another perturbation attack with an additional $k$-NN distance constraint other than Chamfer distance and additional clip operations based on $L_{\infty}$ norms and normal vectors. 
GeoA3 takes Chamfer distance, Hausdorff distance and local curvature loss into account when performing a perturbation attack. 
SI uses specific designs to restrict the direction of perturbations on points. 
HiT identifies attack regions through a two-step process that evaluates saliency and imperceptibility scores, then applies deformation perturbations within these identified regions using Gaussian kernel functions.
The optimization of the perturbations concerning all the attacks above is based on the C\&W attack \cite{carlini2017towards} except for IFGM. 

We provide the hyper-parameter settings for the attacks as follows for potential reproduction needs. 
For all attacks, we first apply the settings from the original works if provided and then tune them for better attack performance on a small validation set. 
For IFGM, we set the attack step size as $0.02$, the number of iterations as $100$, and the clip threshold $\epsilon$ as $0.5$. 
For G3-Pert, we set the attack step size as $0.005$, the number of iterations as $200$, and the distance loss weight as $0.5$. 
For G3-Add, we set the number of points added as $256$, the attack step size as $0.005$, the number of iterations as $400$, and the distance loss weight as $400.0$. 
For AdvPC, we set the number of iterations as $100$, and the distance constraint weight as $0.1$. 
For Drop, we set the total number of points to delete as $200$, and the number of iterations as $20$. 
For KNN, we set the attack step size as $0.001$, the number of iterations as $200$, the $\kappa$ used in the C\&W attack as $5.0$, the Chamfer distance weight as $10.0$, the $k$-NN distance weight as $0.5$, and the clip threshold is $0.03$. 
For GeoA3, we set the max binary search steps of attack step size as $3$, the number of iterations as $50$, and the weights of Chamfer distance, Hausdorff distance, and curvature loss are $10.0$, $0.1$, $1.0$, respectively. 
For SI, we set the attack step size as $0.003$, the number of iterations as $100$, and the clip threshold as $0.08$. 
For HiT, we set the attack step size as $0.005$, the number of iterations as $200$, the $\kappa$ used in the C\&W attack as $30.0$, and the clip threshold $\epsilon$ as $0.2$. The number of candidate center points, local points, and ultimate center points are set as $200$, $32$, and $100$, respectively.
All the attacks are performed with the settings above unless otherwise specified.

\noindent\textbf{Defense Methods. }
We consider three \textit{input preprocessing} methods: SOR \cite{rusu2008towards}, DUP-Net \cite{zhou2019dupnet}, and PointDP (Diffusion) \cite{sun2023critical} as our defense baselines. 
SOR computes the $k$-NN distance for each point in a point cloud and removes those points with distances larger than $\mu + \alpha \cdot \sigma$, where $\mu$ and $\sigma$ denote the mean and standard deviation of the distances. 
DUP-Net further utilizes an up-sampling network \cite{yu2018pu} to enhance the visual quality of a point cloud after SOR. For the hyper-parameters, we set $k$ as $2$ and $\alpha$ as $1.1$ as described in the original work. 
Diffusion leverages a conditional diffusion model to purify the adversarial perturbations within a point cloud with the diffusion and reverse process. We take $5$ steps for each stage, respectively. 

We also consider four \textit{self-robust} models: AT \cite{liu2019extending}, PAGN \cite{liang2022pagn}, PointCutMix(K) \cite{zhang2022pointcutmixa}, and CAP \cite{ding2023cap}. 
For AT and PAGN, we combined adversarial examples generated by the IFGM attack with benign samples for fine-tuning the vanilla classification model. 
For PointCutMix, we augment the training data by replacing a certain proportion of points in one sample with those from another random sample, and then proceed with standard training.
For CAP, we perform contrastive learning within each training batch of samples. 

While we consider these defenses as baselines of our proposed framework, we would like to point out that our framework enhances the inherent robustness of current models, which is compatible with all the existing input preprocessing defense methods, e.g., one can first utilize preprocessing methods to restore an adversarial example before inputting it into our model to further boost the robustness.

\begin{table}[t]
    \caption{Mean test accuracy on ModelNet40 and ShapeNet. }
    \label{table:clean_acc}
    \centering
    \resizebox{0.85\linewidth}{!}{
    \begin{tabular}{ccccc}
    \toprule
    Dataset & Model & PointNet & DGCNN & PCT \\ 
    \midrule
    \multicolumn{1}{c}{\multirow{2}{*}{ModelNet40}} & \multicolumn{1}{c}{Vanilla} & 86.2\%     & 88.9\%  & 89.6\%     \\
    \multicolumn{1}{c}{}                            & \multicolumn{1}{c}{Ours}    & 87.1\%     & 90.9\%  & 91.0\%     \\
    \midrule
    \multicolumn{1}{c}{\multirow{2}{*}{ShapeNet}}   & \multicolumn{1}{c}{Vanilla} & 78.6\%     & 80.5\%  & 77.5\%     \\
    \multicolumn{1}{c}{}                            & \multicolumn{1}{c}{Ours}    & 74.1\%     & 82.0\%  & 76.0\%     \\ 
    \bottomrule
    \end{tabular}
    }
\end{table}

\begin{table*}[t]
    \caption{Classification accuracy (\%) with defenses (\colorbox{TableGreen!20}{input preprocessing$^{\dag}$}, \colorbox{TableBlue1!10}{self-robust$^{\circ}$}) on ShapeNet. Best values are in bold.}
    \label{table:untarget_shapenet}
    \setlength{\tabcolsep}{2pt}
    \resizebox{\linewidth}{!}{
    \begin{tabular}{c@{\hspace{8pt}} cccccc@{\hspace{8pt}} cccccc@{\hspace{8pt}} cccccc}
    \toprule
	\multicolumn{1}{c}{\multirow{2}{*}{Defense}} & \multicolumn{6}{c}{PointNet}		& \multicolumn{6}{c}{DGCNN}		& \multicolumn{6}{c}{PCT}\\
	\cmidrule(r{8pt}){2-7} \cmidrule(r{8pt}){8-13} \cmidrule(r){14-19}
	& IFGM 	& G3-Pert 	& G3-Add 	& Drop 	& AdvPC 	& Avg. 	& IFGM 	& G3-Pert 	& G3-Add 	& Drop 	& AdvPC 	& Avg. 	& IFGM 	& G3-Pert 	& G3-Add 	& Drop 	& AdvPC 	& Avg. \\ 
	\midrule
    \rowcolor{TableYellow!25}
	Vanilla    	 & 0.3	 & 40.8	 & 27.4	 & 57.4	 & 1.9	 & 25.6	 & 0.0	 & 18.5	 & 21.0	 & 68.3	 & 4.8	 & 22.5	 & 22.9	 & 23.1	 & 14.0	 & 66.1	 & 5.1	 & 26.2\\
	\rowcolor{TableGreen!20} SOR$^{\dag}$        	 & 37.1	 & 40.3	 & 31.8	 & 64.2	 & 10.6	 & 36.8	 & 31.3	 & 20.2	 & 22.8	 & \textbf{76.7}	 & 20.2	 & 34.2	 & 46.7	 & 26.2	 & 25.3	 & 65.3	 & 8.6	 & 34.4\\
	\rowcolor{TableGreen!20} DUP-Net$^{\dag}$    	 & 23.9	 & 40.3	 & 5.9	 & 10.0	 & 11.4	 & 18.3	 & 46.9	 & 37.1	 & 4.3	 & 4.1	 & 5.7	 & 19.6	 & 54.2	 & 35.0	 & 30.4	 & 67.2	 & 14.7	 & 40.3\\
	\rowcolor{TableGreen!20} Diffusion$^{\dag}$  	 & 43.7	 & 65.3	 & 27.5	 & 60.8	 & 9.8	 & 41.4	 & 68.0	 & 28.9	 & 22.6	 & 67.2	 & 40.6	 & 45.5	 & \textbf{56.4}	 & 33.2	 & 42.2	 & \textbf{67.8}	 & 12.4	 & 42.4\\
	\rowcolor{TableBlue1!10} AT$^{\circ}$         	 & 23.7	 & 36.9	 & 7.8	 & 24.5	 & 6.6	 & 19.9	 & 25.5	 & 11.7	 & 12.3	 & 37.9	 & 1.9	 & 17.9	 & 11.6	 & 55.1	 & 41.4	 & 58.3	 & 1.0	 & 33.5\\
	\rowcolor{TableBlue1!10} PAGN$^{\circ}$       	 & 32.0	 & 41.4	 & 36.3	 & 63.6	 & 17.7	 & 38.2	 & 14.8	 & 24.7	 & 21.4	 & 70.4	 & 21.0	 & 30.4	 & 43.8	 & 42.4	 & 44.9	 & 60.0	 & 21.2	 & 42.4\\
	\rowcolor{TableBlue1!10} PointCutMix$^{\circ}$	 & 34.4	 & 40.5	 & 32.9	 & 51.1	 & 11.4	 & 34.1	 & 35.1	 & 25.8	 & 32.0	 & 42.6	 & 14.4	 & 30.0	 & 50.3	 & 44.0	 & 38.9	 & 57.2	 & 46.6	 & 47.4\\
	\rowcolor{TableBlue1!10} CAP$^{\circ}$        	 & 33.6	 & 52.8	 & 41.8	 & 55.7	 & 18.6	 & 40.5	 & 4.2	 & 41.8	 & 23.9	 & 60.8	 & 12.6	 & 28.7	 & 41.9	 & 40.4	 & 22.0	 & 62.5	 & 3.5	 & 34.1\\
	\rowcolor{TableBlue1!25} Ours$^{\circ}$       	 & \textbf{69.1}	 & \textbf{69.1}	 & \textbf{62.3}	 & \textbf{69.4}	 & \textbf{68.5}	 & \textbf{67.7}	 & \textbf{72.2}	 & \textbf{72.2}	 & \textbf{56.4}	 & 73.1	 & \textbf{67.6}	 & \textbf{68.3}	 & 56.0	 & \textbf{57.9}	 & \textbf{54.1}	 & 64.0	 & \textbf{59.1}	 & \textbf{58.2}\\
\bottomrule
    \end{tabular}
    }
\end{table*}

\noindent\textbf{Metrics. }
For the untargeted attack, the attack is set to aim at misleading the prediction of the model without any designated label, i.e., causing the model to predict any label other than the ground truth one. 
Therefore, we utilize classification accuracy to measure the effectiveness of an attack, where a higher accuracy indicates better adversarial robustness. 

We use \textit{Silhouette Coefficient (SC)} to measure the quality of the feature space of the models. 
SC is a metric that measures the separation distance between the resulting feature clusters. For a given sample \( i \), let \( a(i) \) denote the mean intra-class distance (i.e., the average distance between \( i \) and all other samples in the same class), and let \( b(i) \) denote the lowest mean inter-class distance (i.e., the average distance between \( i \) and all samples in the nearest different class). The silhouette coefficient \( s(i) \) of sample \( i \) is defined as:
\[
s(i) = \frac{b(i) - a(i)}{\max\{a(i), b(i)\}}
\]
The overall silhouette coefficient for a feature set is the mean of \( s(i) \) over all samples:
\[
\text{SC} = \frac{1}{N} \sum_{i=1}^{N} s(i)
\]
where \( N \) is the total number of samples. The value of SC ranges from $-1$ to $1$, where a value close to $1$ indicates that the samples are well clustered and separated, while a value close to $-1$ indicates that the samples are poorly clustered, i.e., a badly-separated feature space.

\noindent\textbf{Implementation Details. }
For the implementation of the classification models and baseline methods, we directly apply those with released PyTorch implementations. 
For those without source code released or with TensorFlow implementations only, we implement them according to their work. We trained all the models ourselves including those used for classification and those for defense baselines, e.g. DUP-Net \cite{zhou2019dupnet}. 

We consider an untargeted setting for both normal attacks and shape-invariant attacks. 
For attack methods that are originally designed for performing targeted attacks, e.g., GeoA3, we alter their classification loss to an untargeted one, i.e., from minimizing the loss against the target label to maximizing the loss against the ground truth label. 
Further, we may follow the design in the C\&W-based attack \cite{carlini2017towards} and use the margin logit loss. 

All the attacks and defenses are performed on the test set of both datasets. 
Specifically, for time efficiency, we perform GeoA3 \cite{wen2020geometry} attack only on a subset with $800$ samples consisting of $20$ samples from each of the $40$ classes from ModelNet40. 
Note that in the original work, the authors sampled a subset of $250$ pairs of samples and target labels in their experiments as well. 
We believe that our sampling strategy can provide us with a subset of samples representative enough. 

All the experiments are conducted on a machine with a 32-core CPU, 128 GBs of memory, and $2$ NVIDIA 2080 Ti GPUs. 

\section{Illustration of Adversarial Attacks}

\begin{figure}[t]
    \centering
    \includegraphics[width=0.9\linewidth]{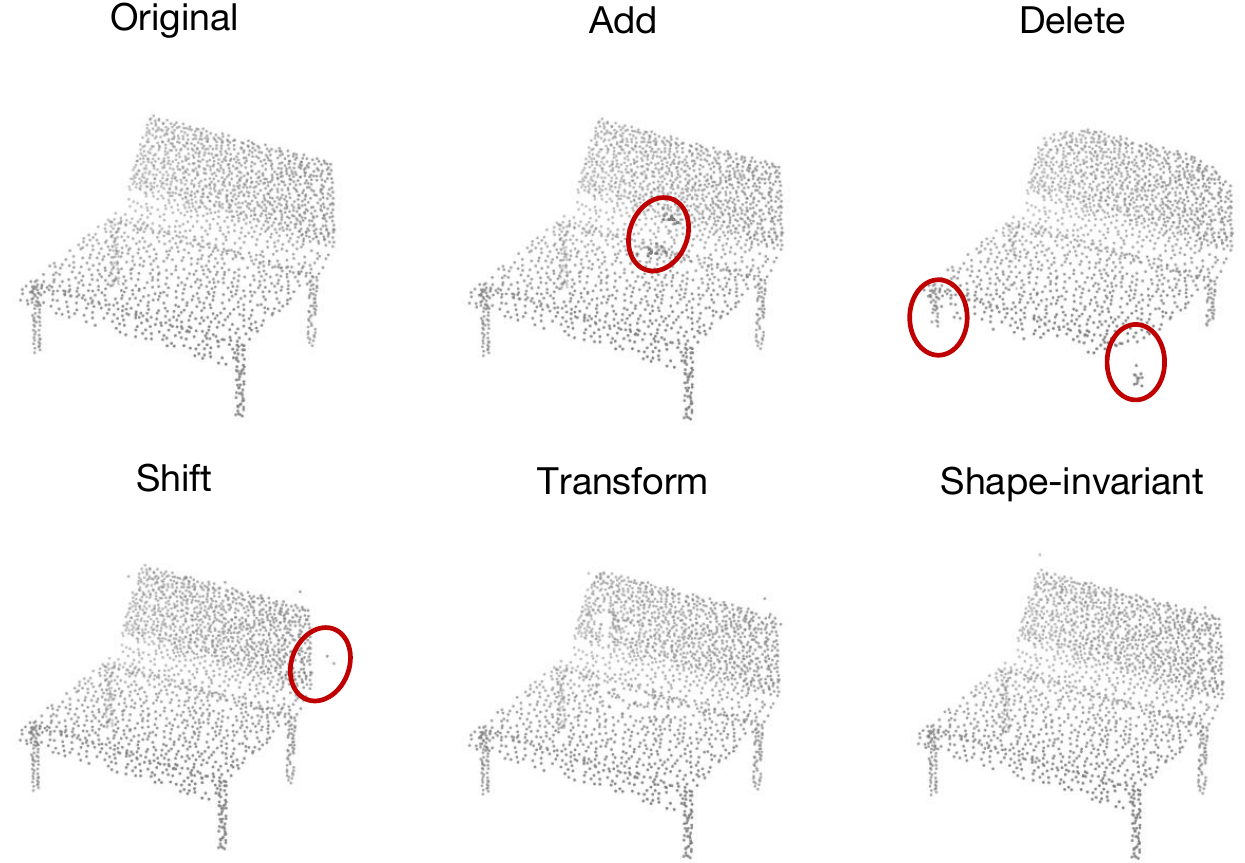}
    \caption{The demonstration of adversarial point clouds generated by various adversarial attacks. }
    \label{fig:intro}
\end{figure}

Fig. \ref{fig:intro} illustrates typical adversarial attacks in point cloud classification. 
Attackers can perturb a point cloud in various ways to mislead the classification model. 
Options include adding \cite{xiang2019generating}, deleting \cite{zheng2019pointcloudsaliency}, or shifting \cite{kim2021minimal} points (i.e., altering the positions of existing points) within a point cloud. 
Also, generative models are used to transform benign point clouds into adversarial ones \cite{hamdi2020advpc,zhou2020lggan}. 
Shape-invariant attacks \cite{huang2022shape,lou2024hide} preserve the geometric structure of benign samples, thereby generating adversarial examples hard to detect.

\begin{table}[t]
    \caption{Classification accuracy (\%) under shape-invariant attacks against various defenses on ModelNet40, evaluated with DGCNN. Best values are in bold, second best values are underlined.}
    \label{table:untarget_shape_invar_dgcnn}
    \setlength{\tabcolsep}{2pt}
    \resizebox{\linewidth}{!}{
    \begin{tabular}{c >{\columncolor{TableYellow!25}}c >{\columncolor{TableGreen!20}}c>{\columncolor{TableGreen!20}}c>{\columncolor{TableGreen!20}}c >{\columncolor{TableBlue1!10}}c>{\columncolor{TableBlue1!10}}c>{\columncolor{TableBlue1!10}}c>{\columncolor{TableBlue1!10}}c >{\columncolor{TableBlue1!25}}c}
    \toprule
	
	Attack & Vanilla	& SOR$^{\dag}$	& DUP-Net$^{\dag}$	& Diffusion$^{\dag}$	& AT$^{\circ}$	& PAGN$^{\circ}$	& PointCutMix$^{\circ}$	& CAP$^{\circ}$	& Ours$^{\circ}$\\ 
	\midrule
	KNN        	 & 12.5	 & 17.5	 & 20.9	 & \underline{47.3}	 & 5.5	 & 9.9	 & 38.3	 & 23.3	 & \textbf{80.4}\\
	GeoA3      	 & 4.8	 & 24.3	 & 16.4	 & \textbf{61.4}	 & 4.5	 & 26.0	 & \underline{53.8}	 & 40.8	 & 31.9\\
	SI	 & 2.5	 & 6.2	 & 8.1	 & 13.5	 & 3.7	 & 2.9	 & \textbf{41.9}	 & 2.5	 & \underline{28.4}\\
	HiT        	 & 70.1	 & 73.8	 & 60.1	 & 72.5	 & 65.6	 & \underline{78.2}	 & 65.6	 & 76.6	 & \textbf{78.5}\\
	Avg.       	 & 22.5	 & 30.5	 & 26.4	 & 48.7	 & 19.8	 & 29.2	 & \underline{49.9}	 & 35.8	 & \textbf{54.8}\\
\bottomrule
    \end{tabular}
    }
\end{table}

\begin{table}[t]
    \caption{Classification accuracy (\%) under shape-invariant attacks against various defenses on ModelNet40, evaluated with PCT. Best values are in bold, second best values are underlined.}
    \label{table:untarget_shape_invar_pct}
    \setlength{\tabcolsep}{2pt}
    \resizebox{\linewidth}{!}{
    \begin{tabular}{c >{\columncolor{TableYellow!25}}c >{\columncolor{TableGreen!20}}c>{\columncolor{TableGreen!20}}c>{\columncolor{TableGreen!20}}c >{\columncolor{TableBlue1!10}}c>{\columncolor{TableBlue1!10}}c>{\columncolor{TableBlue1!10}}c>{\columncolor{TableBlue1!10}}c >{\columncolor{TableBlue1!25}}c}
     \toprule
	Attack & Vanilla	& SOR$^{\dag}$	& DUP-Net$^{\dag}$	& Diffusion$^{\dag}$	& AT$^{\circ}$	& PAGN$^{\circ}$	& PointCutMix$^{\circ}$	& CAP$^{\circ}$	& Ours$^{\circ}$\\ 
	\midrule
	KNN        	 & 46.5	 & 47.1	 & 20.9	 & \underline{47.3}	 & 29.2	 & 35.1	 & 36.4	 & 29.0	 & \textbf{78.9}\\
	GeoA3      	 & 9.8	 & 46.6	 & 52.3	 & \textbf{71.0}	 & 16.3	 & 57.0	 & \underline{59.6}	 & 45.1	 & 50.0\\
	SI         	 & 3.6	 & 13.4	 & 8.1	 & 13.5	 & 5.1	 & \textbf{36.3}	 & \underline{35.7}	 & 22.9	 & 24.6\\
    HiT        	 & \underline{69.4}	 & 59.0	 & 53.5	 & 67.9	 & 64.1	 & 53.2	 & 68.0	 & 54.2	 & \textbf{74.0}\\
	Avg.       	 & 32.3	 & 41.5	 & 33.7	 & 49.9	 & 28.7	 & 45.4	 & \underline{49.9}	 & 37.8	 & \textbf{56.8}\\
\bottomrule
    \end{tabular}
    }
\end{table}

\section{More Empirical Results} \label{sec:more_results}
We present additional experimental results in this section. 

\subsection{Normal Utility}

We begin by evaluating the classification performance of vanilla classifiers enhanced with {\projname}, and present the results on two datasets in Table \ref{table:clean_acc}. 
The results state that the proposed {\projname} offers robustness against adversarial attacks while preserving the model's normal utility. 
Note that existing defenses, whether based on input preprocessing or self-robust models, usually compromise the model’s normal utility to some extent.

\subsection{Robustness against Attacks} \label{sec:more_defense_tables}
We further provide a performance comparison between {\projname} and baseline defense methods on ShapeNet in Table \ref{table:untarget_shapenet}. 
The results align with those of ModelNet40, demonstrating that the proposed {\projname} consistently shows significantly improved robustness across different settings against various types of attacks.

Additionally, we present the results of DGCNN and PCT against shape-invariant attacks on ModelNet40 in Table \ref{table:untarget_shape_invar_dgcnn} and Table \ref{table:untarget_shape_invar_pct}, respectively.
The findings follow a similar pattern to those on PointNet.
We show that existing defenses are effective facing certain attack, such as Diffusion for GeoA3. 
While existing defenses demonstrate effectiveness against specific attacks, such as Diffusion for GeoA3, they fail to withstand diverse attacks with varying patterns.
In contrast, {\projname} achieves the best overall robustness, benefiting from the well-separated feature space facilitated by the ETF head and the adaptive training framework.

\subsection{More Ablation Study} \label{sec:more_ablations}
\begin{table}[t]
	\caption{Classification accuracy (\%) under various attacks against {\projname}
    with different components on ModelNet40, DGCNN/PCT. Best values are in bold, second-best values are underlined. }
	\label{table:ablation_rebuttal}
	\setlength{\tabcolsep}{2pt}
	\centering
	\resizebox{\linewidth}{!}{
	\begin{tabular}{c c c cccccc}
		\toprule                       Model        &   Component            & Clean            & IFGM              & G3-Pert        & G3-Add         & Drop              & AdvPC             & Avg.              \\
		\midrule \multirow{5}{*}{DGCNN}    & Vanilla      & 88.9             & 0.0               & 34.6              & 27.4              & 69.6              & 4.3               & 27.2              \\
		                                            & + ETF        & -2.2             & +77.9             & +44.1             & +34.4             & +7.9              & +73.2             & +47.5             \\
		                                            & + ETF \& RBL & \underline{-0.6} & +55.2             & +33.1             & \underline{+35.7} & +5.2              & +61.1             & +38.0             \\
		                                            & + ETF \& FDL & -2.1             & \underline{+78.9} & \underline{+44.5} & +34.7             & \underline{+10.6} & \underline{+74.2} & \underline{+48.6} \\
		                                            & + All (Ours) & \textbf{+2.0}    & \textbf{+84.3}    & \textbf{+49.6}    & \textbf{+42.5}    & \textbf{+15.0}    & \textbf{+77.0}    & \textbf{+53.7}    \\
		\midrule \multirow{5}{*}{PCT}     & Vanilla      & 89.6             & 7.4               & 64.7              & 56.5              & 82.9              & 26.2              & 47.5              \\
		                                            & + ETF        & +0.9             & +64.6             & +9.8              & +16.0             & \underline{-2.2}  & \underline{+53.2} & +28.3             \\
		                                            & + ETF \& RBL & \underline{+1.1} & +65.9             & \underline{+13.5} & \textbf{17.9}     & -5.1              & +52.8             & +29.0             \\
		                                            & + ETF \& FDL & +0.8             & \textbf{+67.8} & +10.7             & \underline{+17.0} & -3.2              & \textbf{+53.6}    & \underline{+29.2} \\
		                                            & + All (Ours) & \textbf{+1.4}    & \textbf{+67.8}    & \textbf{+16.5}    & +16.1             & \textbf{-0.7}     & +49.4             & \textbf{+29.8}    \\
		\bottomrule
	\end{tabular}
	}
\end{table}

We conduct further ablation experiments for DGCNN and PCT as well on ModelNet40. 
The results are shown in Table \ref{table:ablation_rebuttal}.
The contribution of ETF, RBL and FDL exhibits a similar trend to that in the main paper. 

Moreover, the results of DGCNN and PCT show that models with more advanced architectures can provide a better-aligned feature structure. 
This in turn benefits FDL so that it can improve both the clean performance and robustness with more accurate feature directions.

\begin{table}[t]
    \caption{Inference time cost (ms/sample) of various defense methods. Best values are in bold.}
    \label{table:inference_time}
    \footnotesize
    \centering
    \resizebox{0.85\linewidth}{!}{
    \begin{tabular}{cccccc}
    \toprule
    Model  & Vanilla & DUP-Net & Diffusion & CAP & Ours \\
    \midrule
    PointNet & 0.3 & 0.8 & 2.1 & 4.4 & \textbf{0.2} \\
    DGCNN & 4.2 & 4.7 & 6.0 & 7.5 & \textbf{4.3} \\
    PCT & 8.6 & 9.1 & 10.4 & 11.2 & \textbf{8.8} \\
    \bottomrule
    \end{tabular}
    }
    \vspace{-0.1in}
\end{table}

\subsection{Efficiency}  \label{sec:efficiency}
We evaluate the inference efficiency of defenses. 
Despite the significant robustness improvements achieved by {\projname}, \textit{it introduces only minimal computational overhead compared to vanilla models.}
As shown in Table \ref{table:inference_time}, input preprocessing methods like DUP-Net and Diffusion require additional preprocessing steps, while self-robust models like CAP rely on resource-intensive additional modules.
In contrast, {\projname} simply replaces a model’s classifier with a linear ETF head, keeping the computation cost nearly identical to that of a vanilla model.

\subsection{Visualization of Feature Space} \label{sec:more_defense_visual}

\begin{figure*}[t]
    \centering
    \includegraphics[width=\linewidth]{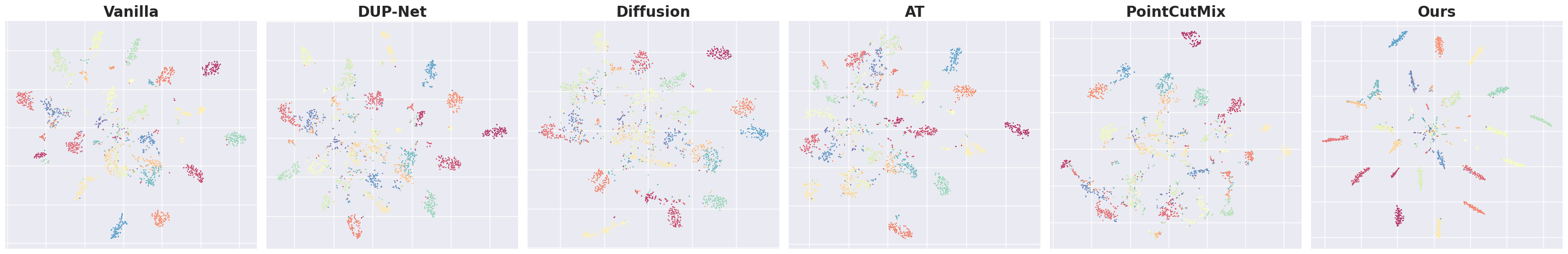}
    \caption{The t-SNE visualization of DGCNN features on ModelNet40 test set under various defense schemes. }
    \label{fig:pilot_tsne_dgcnn}
\end{figure*}
\begin{figure*}[t]
    \centering
    \includegraphics[width=\linewidth]{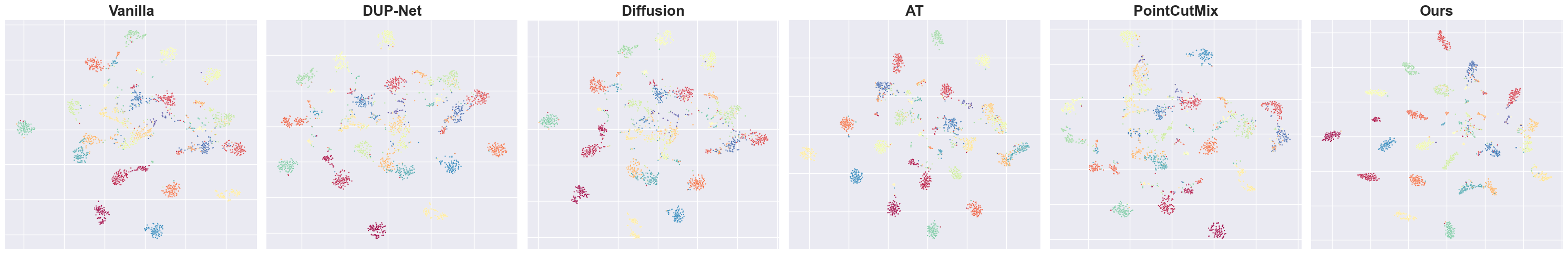}
    \caption{The t-SNE visualization of PCT features on ModelNet40 test set under various defense schemes. }
    \label{fig:pilot_tsne_pct}
\end{figure*}
Similar to Sec. \ref{sec:preliminary_results}, we visualize the feature space of DGCNN and PCT when applying various defenses in Fig. \ref{fig:pilot_tsne_dgcnn} and \ref{fig:pilot_tsne_pct}, respectively.
We show that even advanced models cannot achieve well-disentangled feature space on complex point cloud datasets. 
In contrast, our {\projname} significantly improves the feature separation with the ETF design and the adaptive training framework.

\begin{figure*}[t]
    \centering
    \includegraphics[width=\linewidth]{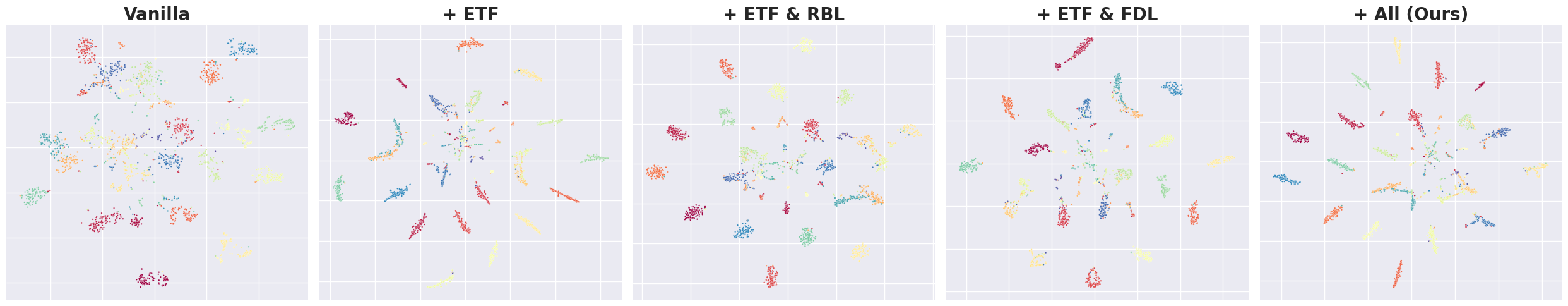}
    \caption{The t-SNE visualization of ablation models based on PointNet, ModelNet40. }
   
    \label{fig:model_tsne}
\end{figure*}
\begin{figure*}[t]
    \centering
    \includegraphics[width=\linewidth]{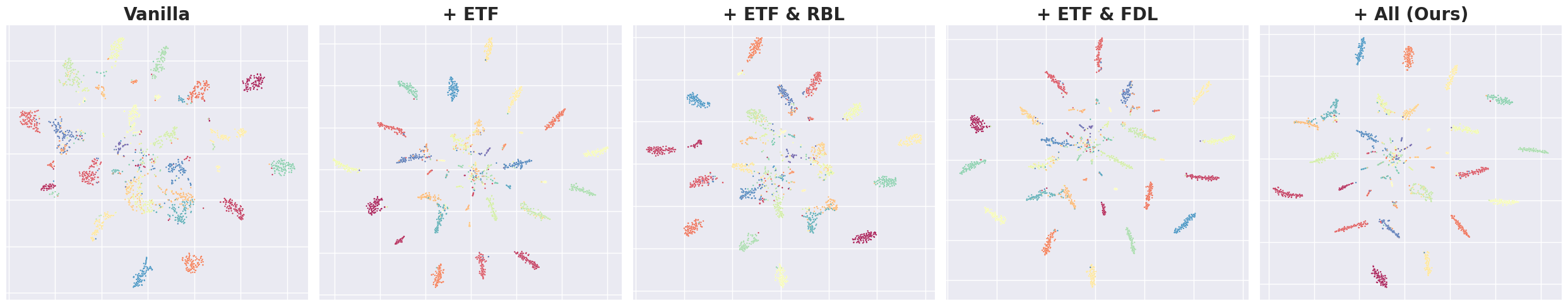}
    \caption{The t-SNE visualization of ablation models based on DGCNN, ModelNet40. }
   
    \label{fig:model_tsne_DGCNN}
\end{figure*}
\begin{figure*}[t]
    \centering
    \includegraphics[width=\linewidth]{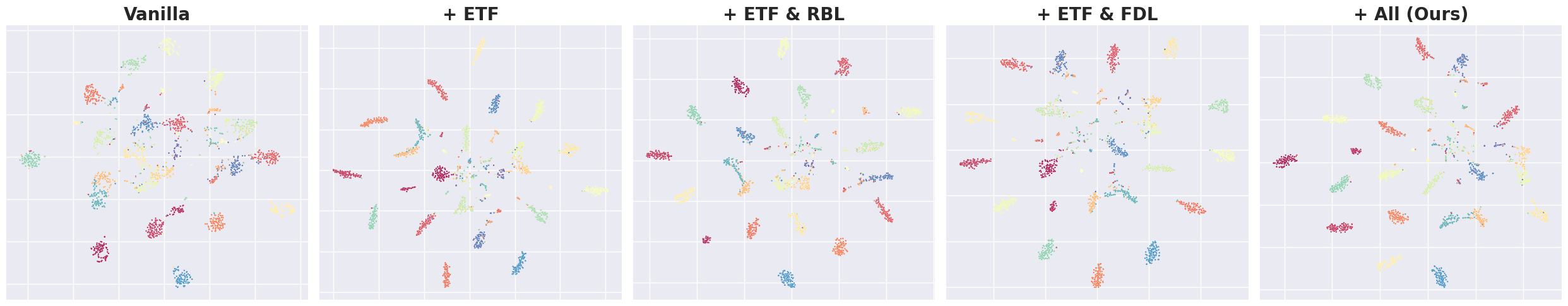}
    \caption{The t-SNE visualization of ablation models based on PCT, ModelNet40. }
   
    \label{fig:model_tsne_PCT}
\end{figure*}

We further present the visualization of the feature space of ablation models on ModelNet40, as shown in Fig. \ref{fig:model_tsne}, \ref{fig:model_tsne_DGCNN} and \ref{fig:model_tsne_PCT}. 
The visualization corroborates our analysis in ablation study. 
A fixed ETF head aligns the feature space of each class with the direction of the classification vector, resulting in strip-like distributions for class features. 
RBL relaxes this alignment, leading to cluster-like distributions with increased class overlap. 
FDL enforces feature direction alignment, making the feature space resemble the strip-like distribution of the fixed ETF. The ideal feature disentanglement is reached by \underline{All}.

\end{document}